\documentclass[10pt, conference, compsocconf]{IEEEtran}
\usepackage{mathptmx}
\usepackage{fancyhdr}
\usepackage[normalem]{ulem}
\usepackage[hyphens]{url}
\usepackage[sort&compress,numbers]{natbib}
\usepackage[final]{microtype}
\usepackage{subeqnarray}
\usepackage{flushend}
\usepackage{bbm}
\usepackage[dvipsnames]{xcolor}
\usepackage{bbding}
\usepackage{relsize}
\usepackage[cmex10]{amsmath}
\usepackage{subfig}
\author{%
\texorpdfstring{
Amir Yazdanbakhsh$^{\ast}$\quad{}\quad{}Ashkan Moradifirouzabadi$^{\ast\dagger}$\quad{}\quad{}Zheng Li$^{\ast\dagger}$\quad{}\quad{}Mingu Kang$^{\dagger}$\\
\smaller{
Google Research, Brain Team\quad{}\quad{}$^{\dagger}$University of California, San Diego\quad{}($^\ast$Equal Contribution) \vspace{0.05cm}}}{Lg}\\
\texorpdfstring{\smaller
{\texttt{\href{mailto:ayazdan@google.com}{ayazdan@google.com}, \href{mailto:ashkan@ucsd.edu}{ashkan@ucsd.edu}, \href{mailto:zhengli@ucsd.edu}{zhengli@ucsd.edu}, \href{mailto:mingu@ucsd.edu}{mingu@ucsd.edu}}}}{Lg}
}

\usepackage{style/hadianeh}
\usepackage{algpseudocode}
\usepackage{tcolorbox}
\usepackage{style/macros}
\usepackage{style/nick_names}
\usepackage{style/results}
\usepackage{tikz}
\usepackage{titlesec}
\usepackage[scaled=0.86]{helvet} 
\definecolor{Gray}{gray}{0.9}


\usepackage[export]{adjustbox}
\usepackage[subtle]{savetrees}
\let\sum\relax
\DeclareSymbolFont{otherlargesymbols}{OMX}{cmex}{m}{n}
\DeclareMathSymbol{\sum}{\mathop}{otherlargesymbols}{"50}
\fancypagestyle{firstpage}{
  \fancyhf{}
  
  \fancyhead[C]{Appears in the Proceedings of the 55th IEEE/ACM International Symposium on Microarchitecture, 2022}
  \fancyfoot[C]{\thepage}
}
\title{Sparse Attention Acceleration with Synergistic\\In-Memory Pruning and On-Chip Recomputation}
\titlespacing{\section}{0pt}{2ex}{1ex}
\titlespacing{\subsection}{0pt}{1ex}{0ex}
\titlespacing{\subsubsection}{0pt}{0.5ex}{0ex}
\usepackage[bookmarks=true,breaklinks=true,colorlinks,citecolor=blue,linkcolor=blue,urlcolor=blue,pdfview=Fit]{hyperref}
\makeatletter
\def\footnoterule{\kern-3\p@
  \hrule \@width 2in \kern 2.6\p@}
\makeatother
\begin{document}
\maketitle
\thispagestyle{firstpage}
\pagestyle{plain}
\begin{abstract}
As its core computation, a self-attention mechanism gauges pairwise correlations across the entire input sequence.
Despite favorable performance, calculating pairwise correlations is prohibitively costly.
While recent work has shown the benefits of runtime pruning of elements with low attention scores, the quadratic complexity of self-attention mechanisms and their on-chip memory capacity demands are overlooked.
This work addresses these constraints by architecting an accelerator, called \sys\footnote{\textbf{\sys}: \textbf{SP}arse attention acceleration with app\textbf{R}oximate \textbf{IN}-memory \textbf{T}oken pruning}, which leverages the inherent parallelism of ReRAM crossbar arrays to compute attention scores in an approximate manner.
Our design prunes the low attention scores using a lightweight analog thresholding circuitry within ReRAM, enabling \sys to fetch only a small subset of relevant data to on-chip memory.
To mitigate potential negative repercussions for model accuracy, \sys re-computes the attention scores for the few fetched data in digital.
The combined in-memory pruning and on-chip recompute of the relevant attention scores enables \sys to transform quadratic complexity to a merely linear one.
In addition, we identify and leverage a dynamic spatial locality between the adjacent attention operations even after pruning, which eliminates costly yet redundant data fetches.
We evaluate our proposed technique on a wide range of state-of-the-art transformer models.
On average, \sys yields \SpeedupOverBaselineS~speedup and \EnergyOverBaselineS~energy reduction when total 16KB on-chip memory is used, while virtually on par with iso-accuracy of the baseline models (on average \AverageAccuracyDegradation\,degradation).
\end{abstract}
\section{Introduction}
\label{sec:intro}
The sweeping success of self-attention mechanisms shifted the focus of our community from Convolutional Neural Networks~\cite{ganax:isca:2018,tpu:isca:2017,bitfusion:isca18,diannao:asplos:2014} to seeking software~\cite{reformer:2020,learnedtoken:2021,long_range_arena,bigbird:2020,gbert:naacl:2019,flat:arxiv:2021} and hardware approaches~\cite{spatten:hpca21,a3:hpca20,leopard:isca:2022,sanger:micro21} to improve efficiency of the attention mechanism.
At its crux, it creates and employs three abstractions of its inputs (e.g. words or pixel patches): query, key, and value embeddings.
The core operation of self-attention is the computation of pairwise correlations between query and key embeddings, followed by computing a weighted sum of value vectors proportional to measured correlations.
Despite its compelling performance, the associated compute and memory footprint cost of self-attention mechanisms can readily become inordinate\footnote{The cost of pairwise correlations grows in the order of $\mathcal{O}(N^2)$ with respect to input sequence length.}, especially as the input sequence length increases (e.g. > 2K), a prevailing trend in recent deep learning models~\cite{long_range_arena,gpt2,compressive_transformer,performer,reformer:2020,liu2018generating}.
To address this challenge, a recent line of research~\cite{a3:hpca20,spatten:hpca21,elsa:isca21,edgebert:micro21,leopard:isca:2022} intuits that \textit{each query is germane to only a dynamic subset of the few key embeddings when determining the input context.} 
This pruning approach appears beneficial, yet does not effectively address the main cost driver of the self-attention mechanism: \textit{data communication overhead}.
This is because identifying the relevance of key embeddings per query, especially to preserve model accuracy, still requires fetching all embeddings to on-chip resources and performing costly query$-$key computations.
Commonly, these methods presume sufficiently large on-chip resources to keep all embeddings for a single head on chip.
This assumption can readily fail, particularly in models with ever-increasing input sequences and in resource-constrained devices.
For example, if we embrace a design with only 20$\%$ of requisite on-chip buffers available for embeddings in a head, data communication emerges as the main determinant of efficiency (on average, $>$\,60$\%$ of total energy consumption as shown in Figure~\ref{fig:dram_energy}).
To address this, we propose in-memory pruning solutions that obviate the need to bring embeddings onto the chip. 

An emerging body of work has illustrated significant benefits of ReRAM in-memory computing, due to the inherent efficiency of analog computing and massive parallelism capability~\cite{reramsearch:ieee:2019,9586231,9499856,9643573,song2018graphr,prime:isca:2016,timely:2020,reramsurvey:ml:2018,pipelayer:hpca:2017}.
We leverage ReRAM technology to enable in-memory pruning, reducing the pressure on the accelerator to fetch all embeddings onto the chip.
While appealing, materializing the possibility of in-memory pruning comes with its own challenges, listed as follows:
\begin{enumpacked}
\item \niparagraph{Circuit inaccuracies:} 
There are various inaccuracies, such as thermal noise, coupling noise, and process variations associated with ReRAM analog circuitry, which limit the precision of in-memory computing.
\item \niparagraph{Data conversion overhead:}
Runtime pruning~\cite{leopard:isca:2022}, a common approach to preserve model accuracy, requires layer-wise comparisons with a threshold value.
The cost of converting the analog results of in-memory computing (multiple bits) to the digital domain for perpetual comparisons against threshold values can outweigh the benefits of in-memory computing. 
\item \niparagraph{Selective read of unpruned embeddings:}
Supporting in-memory ReRAM pruning enforces a particular data layout for key embeddings.
However, this layout constraints the ability to selectively read the unpruned vectors.
\end{enumpacked}

\noindent{}To remedy these considerations, this work makes the following contributions:

\noindent{}\ballnumber{1}
We introduce a unique perspective on the ReRAM in-memory computing paradigm.
We employ approximate in-memory compute and precise on-chip recompute in tandem to mitigate the likely negative repercussions to model accuracy due to inherent circuit inaccuracies. 

\noindent{}\ballnumber{2}
We employ analog comparators to carry out the comparisons with threshold values and instead produce 1-bit data to indicate the pruning status.
With this shift in design, we reduce the hardware cost, which is proportional to input bit precision, to merely the cost of a series of 1-bit analog to digital converters (ADCs). 

\noindent{}\ballnumber{3}
We repurpose an existing solution, which enables us to implement data reuse based on our observations.
On the hardware side, we rely on recently taped-out transposable ReRAMs~\cite{trans:isscc:2020} that introduce in-situ transposed read access.
While initially intended for efficiently accessing neural network weights, our application of this hardware selectively reads unpruned embeddings.
For the data reuse, we observe that there is a considerable spatial locality between unpruned key vectors of adjacent queries. 
We exploit this spatial location to improve data reuse and further reduce the data communication overhead.

We evaluate our approach in several self-attention models with large sequences, including \bench{BERT}, \bench{ALBERT}, \bench{ViT}, \bench{GPT-2}, and two futuristic designs (e.g. 2K and 4K input sequence length).
Under an iso design, our results show that, on average, \sys delivers \SpeedupOverBaselineS  ~speed-up and \EnergyOverBaselineS ~energy reduction compared to a baseline design with 16KB on-chip memory.
The benefit increases as on-chip resources become scarcer, representing a design point for resource constrained platforms, e.g. 1.6$\times$ more energy reduction with 16KB on-chip memory than the case with 64KB capacity.
\section{Background and Motivation}
\label{sec:overview}
\subsection{Background}
\niparagraph{Self-attention computations.}
\label{subsec:overview:self-attention}
``\textit{Self-attention}'' computes pairwise correlations across the entire input sequence~\cite{vaswani2017attention}.
Each input element, a word or a pixel patch, is encoded to a vector of size $1\times{e}$. 
We then project these embeddings onto three latent spaces by multiplying each vector into distinct learned weight matrices, $W^\mathcal{Q}$, $W^\mathcal{K}$, and $W^\mathcal{V}$ for \rev{query, key,} and value vectors, respectively.
In the next step, we calculate $s$ scores per query vector $q_i$ ($i$-th row out of $s$ rows in $\mathcal{Q}$), each score representing the relevance of a query to all the key vectors, including itself.
The resulting scores are then normalized by employing a row-wise ``\textit{Softmax}'' operation, producing an attention probability matrix.
A higher probability value indicates greater relevance of the corresponding element with respect to others in the input sequence.

Finally, to obtain the attention values ($\mathcal{A}_{s\times{d}}$\rev{)}, the probability matrix is multiplied by the value matrix followed by a sum reduction.
Intuitively, the objective of this step is to scratch out the value of the low probability elements, while intensifying the others.
To further improve the performance of the self-attention mechanism, multiple paths with dedicated query, key, and value weigh matrices are introduced.
This self-attention is generally known as ``multi-headed''.
Under this paradigm, final values are generated via a concatenated form of attention vectors from each head, which is projected onto a matrix of size $s\times{}d_w$ using one or more feed-forward layers.

\niparagraph{Learned runtime pruning.}
The self-attention mechanism is prohibitively costly in terms of computation and memory, in the order of $\mathcal{O}(s^2)$.
Recently proposed methods~\cite{a3:hpca20,leopard:isca:2022,spatten:hpca21} prune attention values with low scores by banking on inherently large redundancy in input sequences.
While~\cite{a3:hpca20,spatten:hpca21} trades model accuracy for higher performance, \leopard~\cite{leopard:isca:2022} proposes a learned runtime pruning method trailed by an early compute termination mechanism to ensure on par model accuracy with baseline.
Once complete, it incorporates the learned threshold values during inference to prune inconsequential scores, after performing the entire computation of $\mathcal{Q}\times{\mathcal{K}^T}$, cutting down most of the computations after Softmax.
A common theme among existing methods is the assumption of sufficiently large on-chip buffers to store the entire key and value matrices.
However, this assumption fails at longer input sequences~\cite{long_range_arena,compressive_transformer,performer,reformer:2020,liu2018generating} (e.g. > $2K$) as well as for resource-constrained accelerators.
This work builds upon the learned runtime pruning method introduced in~\cite{leopard:isca:2022, learnedtoken:2021} and specifically tackles the pressures on on-chip capacity and data communication overhead. 
Our objective is to eliminate unnecessary data communications and on-chip computations of $\mathcal{Q}\times{\mathcal{K}^T}$ by approximating the thresholding mechanism inside memory.
\begin{figure}[t]
\centering
\includegraphics[width=0.99\columnwidth]{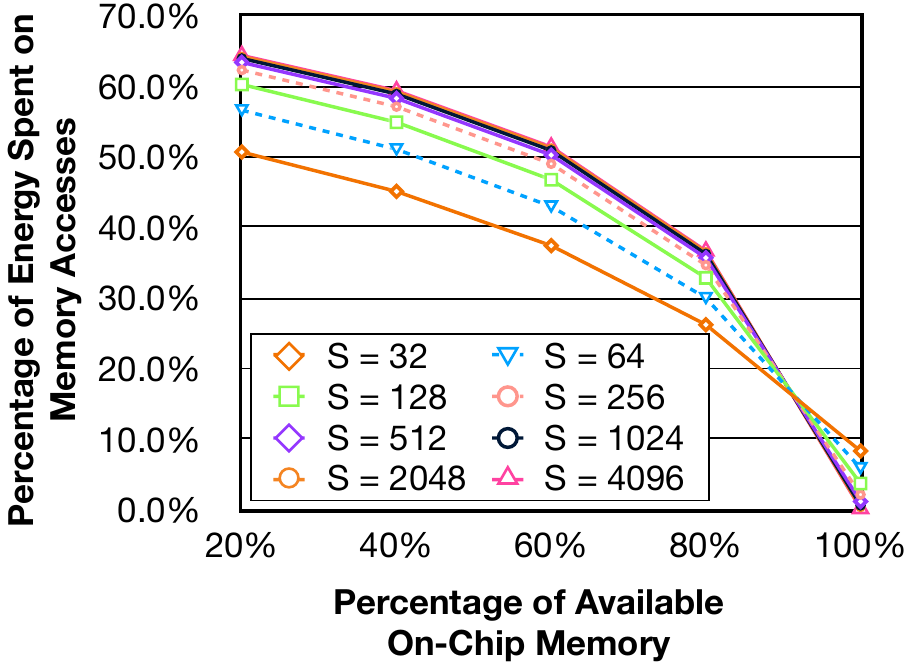}
\caption{Percentage of energy spent on memory accesses to process one attention head with respect to various percentages of available on-chip memory. The results are shown across various sequence ($\mathcal{S}$) length.}
\label{fig:dram_energy} 
\end{figure}
\subsection{Motivation}
\label{subsec:motiv}
As the input sequence length is poised to increase dramatically for future transformer models (e.g. $s=$\,1024 and 2048 in GPT~\cite{gpt2}), motivated by the resulting improvements in model performance~\cite{performer,reformer:2020,liu2018generating}, the assumption of sufficient on-chip resources is no longer valid.
Additionally, as transformer models pave their way into resource-constrained devices~\cite{yazdanbakhsh2021evaluation}, the increased demand for on-chip memory capacity and higher compute efficiency form a challenging design target, even for transformer models with modest sequence length (e.g. $s=$\,[128, 256, 384] for BERT~\cite{gbert:naacl:2019,albert:iclr:2019}).

While recent literature~\cite{a3:hpca20,spatten:hpca21,leopard:isca:2022} favors pruning, these approaches nevertheless are plagued by the considerable overhead of data communication even with adequately sized on-chip buffers.
This cost is exacerbated when on-chip resources are limited, because of frequent instances of data communication.
Figure~\ref{fig:dram_energy} measures the contribution of off-chip memory read and write accesses to the overall energy consumption to process a single-head self-attention layer\footnote{Section~\ref{sec:eval} outlines the experimental setup details.}.
The x-axis encompasses various fractions of on-chip memory capacity with respect to different input sequence lengths.
As on-chip resources become scarce (20$\%$ of requisite on-chip buffers available to store the entire key and value matrices), on average, the energy contribution of on-chip memory increases to $>$\,60$\%$, turning into the dominant energy contributor. 
In this tightly-budgeted scenario, approaches that unlock the opportunity to fetch only a subset of relevant data become attractive.

One such compelling solution is applying the run-time pruning such as  \cite{leopard:isca:2022,a3:hpca20,spatten:hpca21,sanger:micro21}. 
However, even these techniques require to bring in the entire key and value matrices to exercise thresholding.
This research tackles above challenge by approximating $\mathcal{Q}\times{\mathcal{K}}^T$, followed by a comparison with threshold values.
Despite the approximation, our results ensure that this in-memory thresholding mechanism can consistently identify the entire subset of relevant vectors.
To guarantee accuracy on par with baseline, we recompute the score values in a precise manner after selective data fetching.
\subsection{Data Communication Optimization}
\label{data movement reduction opportunity}
Processing self-attention scores with limited on-chip memory capacity requires frequent data movement between adjacent query vectors.
This section points out several opportunities to cut down the cost of such movements.

\subsubsection{In-memory Thresholding}
Under scarce on-chip resources, a logical optimization step can leverage in\rev{-}memory computing to eliminate inconsequential data communications for pruned key and value vectors.
For example, in Figure~\ref{fig:spatial locality}, the core simply stipulates $\mathcal{K}_{2,4,5,6,11,13}$ for $q_1\times{\mathcal{K}^T}$ computations ($q_1\rightarrow$\qk{The}).
This observation provides the opportunity to significantly cut costs by informing the accelerator to only fetch the requisite data.

\subsubsection{Spatial Locality in Adjacent Queries}\label{subsec:spatial}
While in-memory thresholding trims down the amount of data per query that are brought into on-chip buffers, it increases the frequency of data fetches.
This is because a new set of key and value vectors should be fetched to proceed computing for subsequent queries once the computations for $q_i\times\mathcal{K}^T$ completes.
This increase in the frequency of data fetches may well neutralize the potential benefits of reducing the amount of transferred data. 

\begin{figure}[t]
\centering
\includegraphics[width=0.9\columnwidth]{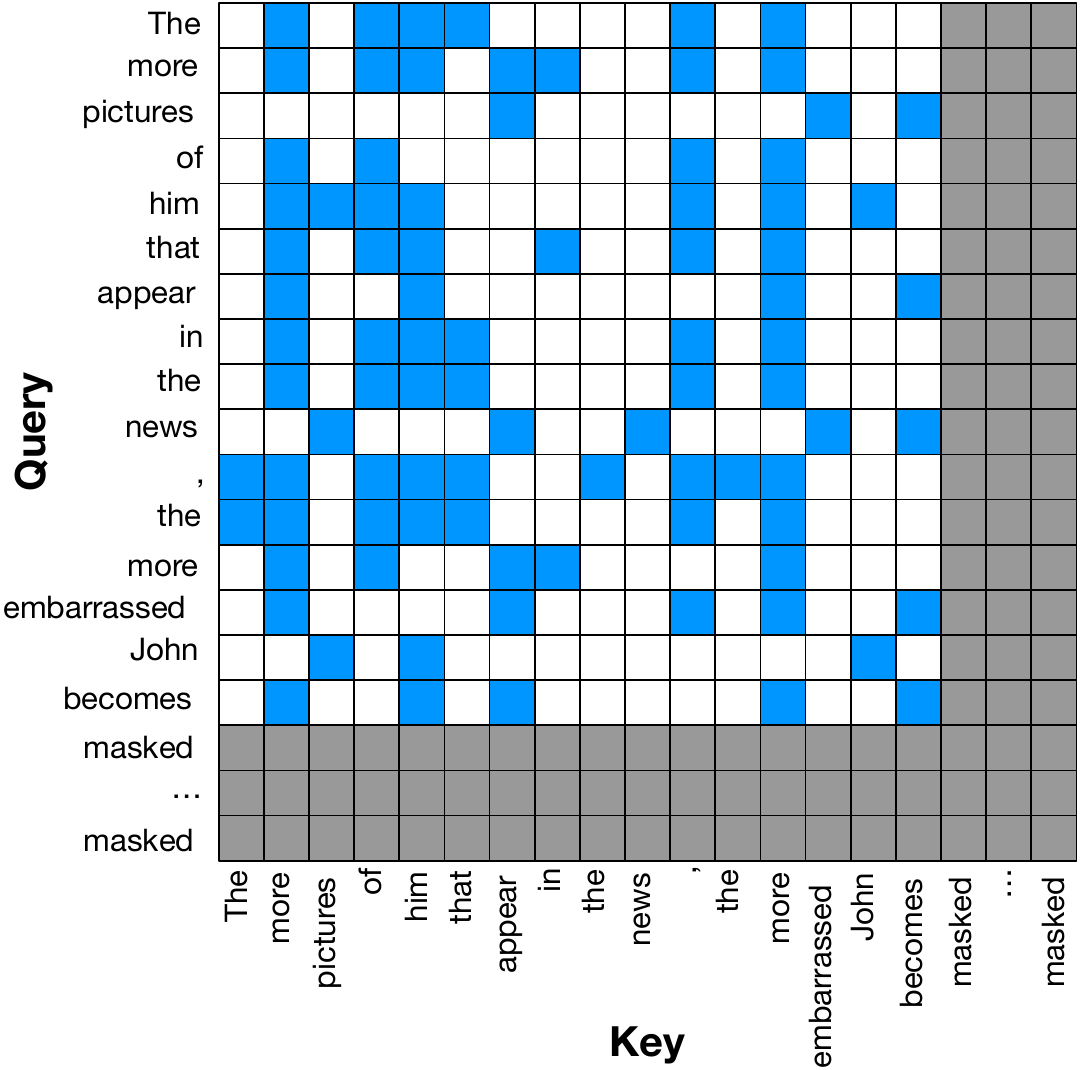}
\caption{Query-Key relation for the first attention layer of CoLA task from GLUE dataset~\cite{glue:arxiv:2018}. White squares represent pruned entries. The gray stripes are masked regions.}
\label{fig:spatial locality} 
\end{figure}

To explore future potential reductions in the amount of transferred data and compensate for the likely overhead of frequent data transfers, we study the similarities between unpruned keys across input queries.
Figure~\ref{fig:spatial locality} illustrates a real example of CoLA task from GLUE dataset~\cite{glue:arxiv:2018} (eighth head in the first attention layer).
Each row indicates a query and its corresponding unpruned key locations, filled in \rev{blue}.
The grey shading on the last few rows and columns specifies the input mask, commonly used in transformer models when the sequence length in the input dataset is less than the one in the model.
It is visually evident that a significant number of keys are inconsequential per query, and that there is a high spatial locality between adjacent rows.
For example, compared to query \qk{The}, the additional required keys for the adjacent query \qk{more} are only \qk{appear} and \qk{in}.
The remaining unpruned key elements, such as \qk{more}, \qk{of}, and \qk{him}, are identical between these queries, obviating additional data transfers.

\niparagraph{Theoretical expectation of spatial locality.}
Equation~\ref{eq:comb} calculates the probability of $\mathcal{L}$, defined as the number of overlapping elements between adjacent queries of size $\mathcal{S}$.
In this equation, $\mathcal{M}$ represents the number of the unpruned elements in each query.
The probability of $\mathcal{L}$ is calculated by first multiplying the numbers of possible combinations of $\mathcal{L}$ elements out of $\mathcal{M}$ and the remaining $\mathcal{M}$ - $\mathcal{L}$ elements out of $\mathcal{S}$ - $\mathcal{M}$.
This product is subsequently divided by the number of possible combinations of $\mathcal{M}$ elements out of $\mathcal{S}$.
The resulting probability of each $\mathcal{L}$ is then multiplied by the value of $\mathcal{L}$ and summed across $\mathcal{M}$ to calculate the theoretical expected overlap between adjacent queries, as demonstrated in Equation \ref{eq:comb}.
\begin{equation}
\label{eq:comb}
\begin{split}
  \mathop{P}(\mathcal{L}) = \frac{\Mycomb[\mathcal{M}]{\mathcal{L}}\times\Mycomb[\mathcal{S}-\mathcal{M}]{\mathcal{M}-\mathcal{L}}}{\Mycomb[\mathcal{S}]{\mathcal{M}}},\hspace{0.2cm}
 \mathop{E}(\mathcal{L}) = \mathlarger{\sum}_{\mathcal{L}=1}^{\mathcal{M}}\mathcal{L}\cdot{\mathop{P}(\mathcal{L})}
\end{split}
\end{equation}
\begin{figure}[t]
\centering
\includegraphics[width=0.99\columnwidth]{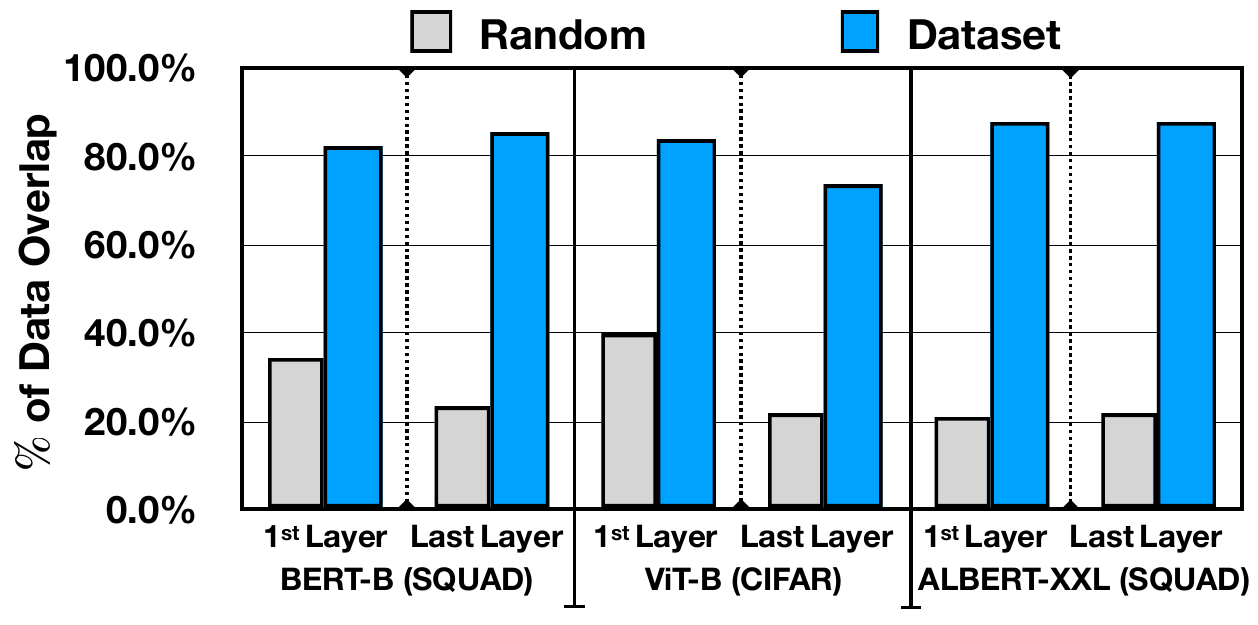}\vspace{-0.2cm}
\caption{Number of common indices between neighboring tokens ($\mathcal{Q}_i$ vs. $\mathcal{Q}_{i+1}$) with the practical dataset vs. randomly selected pruned tokens with the pruning rate from~\cite{leopard:isca:2022}.
}
\label{fig:quantifying_spacial_locality} 
\end{figure}
In Figure \ref{fig:quantifying_spacial_locality}, we compare the percentages of overlaps, averaged across multiple inputs and observed in various extant datasets~\cite{gbert:naacl:2019,vit,gpt2}, with the theoretical expectation formula, as presented in Equation~\ref{eq:comb}.
The results reveals a striking 2 - 3$\times$ increase in the observed overlap percentage in the real world scenarios.
This increase highlights a notable data reuse opportunity because most of the requisite elements already reside in on-chip buffers.
Therefore, exploiting this data reuse opportunity limits the number of data fetches only to the unpruned elements that differ between adjacent queries, leading to a dramatic cost reduction.
\rev{One could leverage spatial locality across larger windows ($>$2) at the cost of hardware complexity. However, on average, the resulting overlap is below 5$\%$, which does not justify the requisite  overhead.}

\subsubsection{Futile Computations in Padded Regions}
\label{sec:padding}
It is a common practice~\cite{huggingface:2019} in transformer models to pad input sequences that are shorter than the maximum supported length.
The padded inputs do not meaningfully contribute to the self-attention computations, and hence are irrelevant for the final model accuracy.
These padded regions are highlighted as gray squares in Figure~\ref{fig:spatial locality}, where only 16 queries out of 128 are computationally relevant. 
This leaves (128-16)$\times$(128-16) score computations inconsequential.
The padded regions are commonly nullified by placing a sufficiently large negative value~\cite{huggingface:2019}.
Passing these negative values through Softmax prompts their probability to approach zero, excluding them from subsequent computations.
To further eliminate unnecessary data communications in these padded regions, we can proactively identify them as early as possible in memory.
\section{In-memory Thresholding}
\label{subsec:inmemBkg}
\begin{figure}[t]
\centering
\includegraphics[width=0.95\columnwidth]{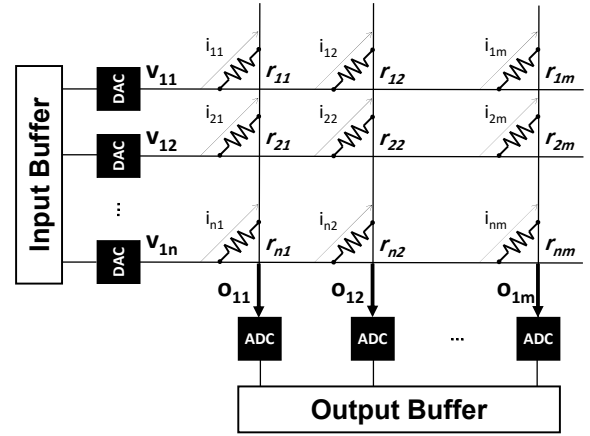}
\caption{In-memory computing with ReRAM cross-bar array.}
\label{fig:reram_in_memory} 
\end{figure}

\niparagraph{Overview of ReRAM.}
Resistive Random Access Memory (ReRAM) is a non-volatile memory that stores data using its adjustable resistance.
Figure~\ref{fig:reram_in_memory} demonstrates a ReRAM 2D crossbar array~\cite{crossbar:2012}.
To further improve the density and energy efficiency of ReRAM, recent methods~\cite{MLC1,MLC2, MLC3,mlc4} use Multi-Level Cells (MLC) to store multiple bits of information inside each cell.
In contrast to Single-Level Cells (SLC), the MLC ReRAM permits a range of resistance values inside each cell.
Although storing more bits per cell appeals by increasing ReRAM memory density, it can easily become a limiting factor.
As the number of bits/cell increases, each cell renders itself more amenable to circuit noises and limits the accuracy of computations.
Recent studies~\cite{1t1mmultiply:2016,fourreram} deem a four bits/cell MLC ReRAM design the optimal balance between robustness and complexity of current sensing detection circuitry.

\niparagraph{Vector-Matrix multiplication with ReRAM in-memory computing.}
ReRAM can perform efficient and highly parallel analog vector-matrix multiplications, as demonstrated by prior work~\cite{pipelayer:hpca:2017,prime:isca:2016,isaac:isca:2016,reramsearch:ieee:2019} on DNN acceleration.
To perform such multiplications, the matrix elements are mapped onto memristor conductance and the input vector is fed into ReRAM's wordlines (Figure~\ref{fig:reram_in_memory}, horizontal lines), one element per row, as biased voltages generated by a digital-to-analog converter (DAC). 
Additionally, a sum reduction can be executed on the resulting multiplications across the crossbar columns as serial currents~\cite{yao2017face,timely:2020}.
Once complete, the weighted-sum vector forms an analog current at the boundary of the ReRAM crossbar, one element per column.
The following equation formally presents a multiplication between vector $v_{1\times{n}}$ and matrix $M_{n\times{m}}$ on a ReRAM crossbar array:
\begin{equation}
\label{eq:reram_cell}
\begin{split}
m_{ij} = \frac{1}{r_{ij}}\hspace{0.5cm}o_{1j} = \mathlarger{\sum}_{i=1}^{n}v_{1i}\cdot{m_{ij}}
\end{split}
\end{equation}
\noindent{}where $m_{ij}$ and $r_{ij}$ represent each element of matrix $M$ and its corresponding resistance value in ReRAM cells.

\niparagraph{Application in run-time pruning.}
The in-memory principle introduced above can be seamlessly applied for accelerating the attention mechanism.
This can be achieved by storing each $k_i$ vector in a column of the crossbar array, and applying the input voltage level, which corresponds to the element of query vector $q_i$, to each wordline as described in Figure~\ref{fig:transposable_rram}(a).
Ideally, we require $s$ columns to store entire sequence length while $d$ rows are needed to accommodate the entire embedding size. If the array size does not match with problem size, multiple banks of array can be employed in a tiled manner.
All of $k_i$ vectors stored in multiple columns are processed for  parallel dot-product operations in one shot. Once it completes, the next query vector $q_{i+1}$ is processed in the subsequent cycle.

\niparagraph{Analog$\leftrightarrow$Digital challenges.}
It is shown~\cite{prime:isca:2016,isaac:isca:2016,timely:2020} that digital$\leftrightarrow$analog conversion drains a significant portion of total ReRAM power consumption, especially as the number of conversion bits increases.
For example, the power and area of a 5-bit ADC are >20$\times$~\cite{murmann2018adc} and >30$\times$~\cite{flash_adc} higher than a 1-bit ADC, respectively.
Therefore, it is crucial to take the power overhead of these converters into account, especially for designs with greater than one-bit precision requirements.
In the following, we discuss the main challenges to in-memory thresholding.

\subsection{In-Memory Thresholding Challenges}
\label{subsec:inmemth}

\noindent\ballnumber{1}\,\niparagraph{Analog computing inaccuracies.}
Analog computing in ReRAM is commonly known to be susceptible to inherent circuit noises and inaccuracies, such as thermal noise, temperature fluctuations, process variations, and coupling noise between adjacent cells~\cite{reramthermal,thermalreram2,reramcircuit}.
These inaccuracies limit the feasible precision of computations in ReRAM crossbar arrays.
\begin{figure}[t]
\centering
\includegraphics[width=0.99\columnwidth]{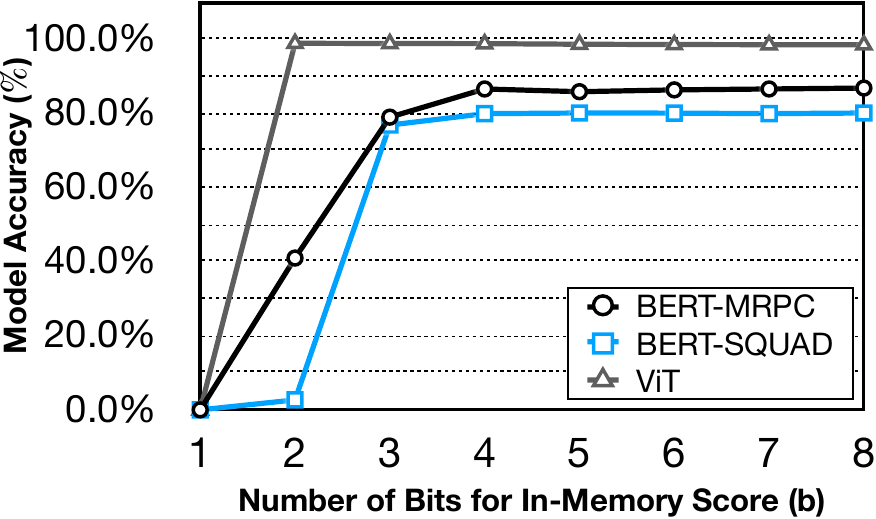}
\caption{Sensitivity of model accuracy to the number of bits ($b$) used for in-memory thresholding (comparison of in-memory scores with $\mathcal{T}h$, Equation~\ref{eq:comp_error}).}
\label{fig:errorresiliency} 
\end{figure}
To evaluate the impact of limited compute precision for in-memory thresholding on the final model accuracy, we use the following approach\footnote{As we explain in Section~\ref{sec:eval}, we do \textit{not} perform additional fine-tuning to quantize key values to lower bit-precision.}:
\begin{equation}
\label{eq:comp_error}
\begin{split}
  \mathcal{P}rune = \mathrm{Argwhere}~(\mathrm{Score_{R}^{b}} < \mathcal{T}h);\hspace{0.2cm}
  \mathrm{Score}[\mathcal{P}rune] = -c
\end{split}
\end{equation}
\noindent{}where $\mathrm{Score_{R}^{b}}$ denotes the in-memory score values (e.g. results of $q_i\times{\mathcal{K}^T}$) when the output has limited accuracy with a $b$-bit precision.
``Argwhere'' finds the indices of score elements that are lower than the target threshold.
Note that the threshold values ($\mathcal{T}h$) are learned during the full-precision finetuning process such as \leopard~\cite{leopard:isca:2022, learnedtoken:2021}. 
The scores of the identified pruning indices are then forcefully set to a large negative value ($-c$) to remove irrelevant elements.
Also, recall that we perform low-precision in-memory computing for the sole purpose of identifying the irrelevant key vectors.
With on-chip accelerators, the score computation for unpruned vectors is still performed in full-precision.

Figure~\ref{fig:errorresiliency} compares the final model accuracy after quantizing the $\mathrm{Score}$ with different bit-precision ($b$) across three different models: BERT-Base~\cite{gbert:naacl:2019} with GLUE~\cite{glue:arxiv:2018}, BERT-Base with SQUAD~\cite{squad:arxiv:2016}, and ViT~\cite{vit} with \cite{cifar10} dataset.
The results show that the quantization error with 4-bit precision virtually has no impact on the final model accuracy\footnote{A recent study from HP Lab~\cite{1t1mmultiply:2016} has shown that ReRAM in-memory computing for 64-tap dot-product delivers 5-bit equivalent output accuracy after including all the error sources.}.
Thus, the runtime pruning mechanism is robust against approximation, even when the computation has a certain level of errors.
This is intuitive because the incorrectly pruned vectors already exhibit a small score value, likely in the vicinity of $\mathcal{T}h$. 
Hence, the impact on model accuracy is negligible.
Finally, even more sensitive workload to the noise can be in theory compensated by adding a modest negative margin on top of $\mathcal{T}h$ in Equation~\ref{eq:comp_error} at the cost of reducing the pruning ratio (directly proportional to hardware performance).

\noindent\ballnumber{2}\,\niparagraph{ADC converter overhead.}
The overhead of ADC converters increases proportionately to the precision of conversion.
Two design choices can support comparisons between vector-matrix multiplication outputs and the threshold values.
The first option uses a 5-bit ADC to convert the outputs and employs digital comparators for thresholding.
The other option utilizes analog comparators for thresholding prior to ADC.
The output of each analog comparison represents a binary value, which indicates whether to prune the corresponding key vector.
Since the resulting pruning vector only requires one bit per key, we can use a low-overhead 1-bit ADC (implemented as a comparator).
The low overhead of 1-bit ADC (>20$\times$~\cite{murmann2018adc} lower area and >30$\times$~\cite{flash_adc} lower power consumption compared to a 5-bit ADC) favors the second option for in-memory thresholding.

\noindent\ballnumber{3}\,\niparagraph{Reading unpruned vectors overhead.}
Finally, performing in-memory thresholding followed by fetching each unpruned $\mathcal{K}$ vector from ReRAM arrays (for digital re-compute) is arduous and can impose significant read latency. 
This occurs because we store each vector of $\mathcal{K}$ vertically at each ReRAM column (Figure~\ref{fig:reram_in_memory}), and $k_{i}$ is mapped to the $i^{th}$ ReRAM column.
On the other hand, accessing from ReRAM through a standard read operations fetches the data stored horizontally in a row.
Therefore, fetching from ReRAM requires sequentially asserting \textit{all} the (horizontal) wordlines, bringing in each row of the $\mathcal{K}$ matrix (even the ones associated with \textit{pruned $k$ vectors}), and selectively fetching the unpruned vectors to on-chip buffers.
We address this challenge by a recent taped-out transposable ReRAM proposal~\cite{trans:isscc:2020}, which we expound below.

\subsection{Transposable ReRAM for Thresholding}
\label{subsec:transthr}
\niparagraph{Overview.}
\begin{figure}[t]
\centering
\includegraphics[width=0.99\columnwidth]{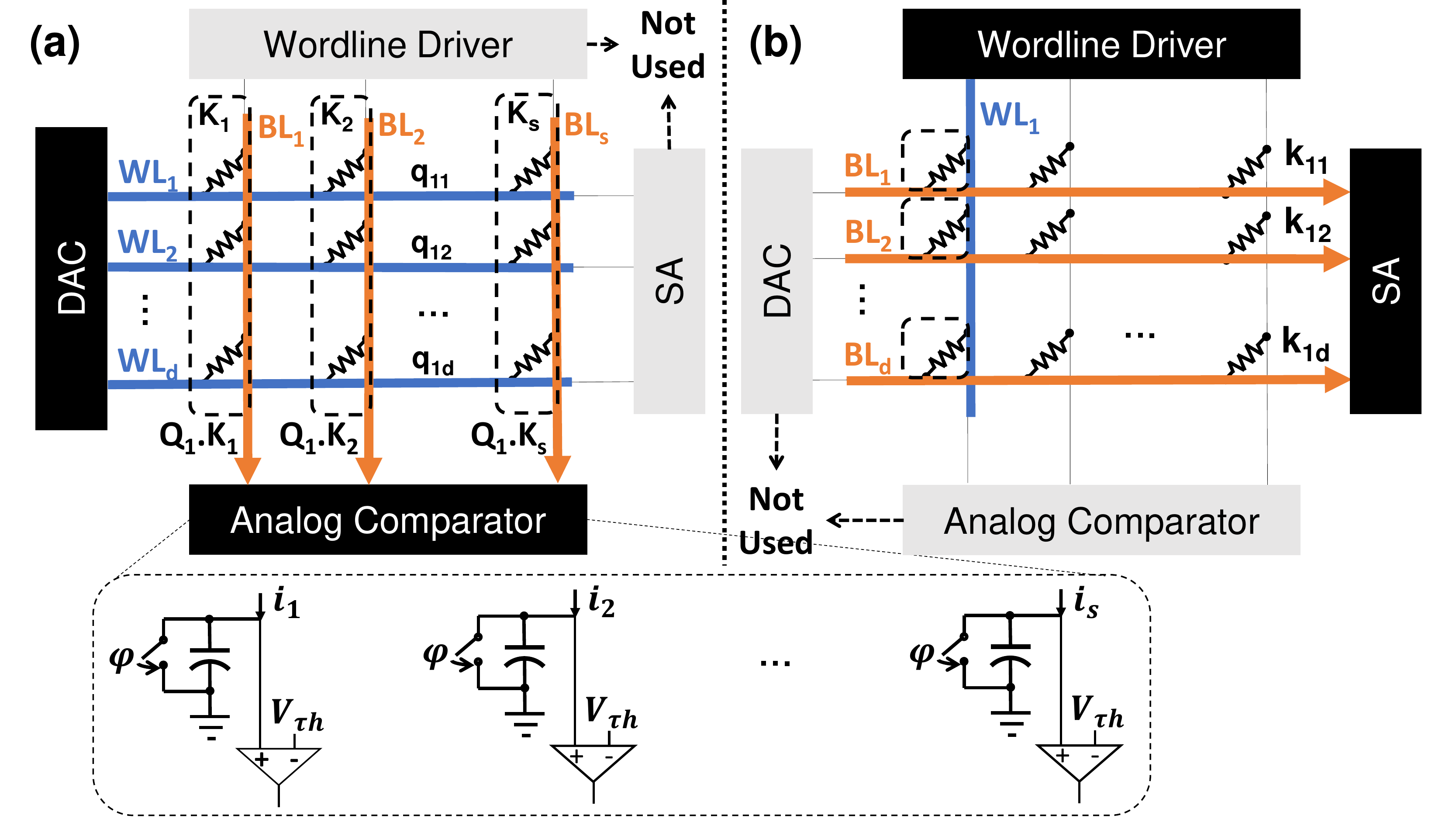}
\caption{Transposable ReRAM crossbar array.  (a) ReRAM crossbar during in-memory pruning, (b) Transposed ReRAM crossbar during normal read.} 
\label{fig:transposable_rram} 
\end{figure}
A transposable ReRAM~\cite{trans:isscc:2020} supports (1) in-situ access to the array to perform vector-matrix computations (in-situ computation), as well as (2) reading \textit{their transposed} values (transposed read).
Figure~\ref{fig:transposable_rram} shows the overall design of a transposable ReRAM in these two modes.
In the ``in-situ computation'' mode, the ReRAM array performs vector-matrix multiplications, similarly to conventional (non-transposable) ReRAM crossbar shown in Figure~\ref{fig:reram_in_memory}.
In this case, we assign the value of each element in input vector $q_i$ to wordlines (horizontal) and assert all the bitlines (vertical) to enable parallel multiplications.
On the other hand, in the new ``transposed read'' mode, the horizontal lines become bitlines and vertical line becomes wordline. In this mode, \textit{only} one wordline gets asserted. 
Once the bitline current from all the columns are fully developed, the sense amplifier reads all the values stored on the ReRAM conductance of the asserted wordline (in the column). 

\niparagraph{In-memory thresholding dataflow.}
As discussed in the previous section, one of the challenges for performing in-memory thresholding is reading unpruned vectors after score calculation (\ballnumber{3}).
The ``transposed read'' mode presents a viable solution to this challenge.
Next, we present a dataflow to identify unpruned key vectors leveraging transposable ReRAMs.
In this dataflow, we store each key vector vertically in the ReRAM crossbar array (the first key vector is mapped onto the first ReRAM column, and so on).
Because analog circuit noises limit the supported bit-precision on each memory cell, we \textit{only} store a predefined subset of MSB-side bits within each cell.
Our experiment showed that a 4-bit precision is sufficient for in-memory thresholding, yielding on par model accuracy.
As such, we only store four MSBs per key vector element in transposable ReRAM arrays.
The rest of LSBs can be stored on conventional ReRAM modules.
Similarly, the elements of query and value vectors are stored on conventional ReRAM modules.
Note that these modules do not need any support for in-memory computations and are solely used for storage\footnote{We homogeneously use ReRAM for storage of queries and values and in-memory thresholding for simplicity. Another possibility can exploit a heterogeneous design, in which DRAM memories are used for query/value matrices and small ReRAM crossbar arrays for in-memory thresholding.}.

\begin{figure}[t]
\centering
\includegraphics[width=0.99\columnwidth]{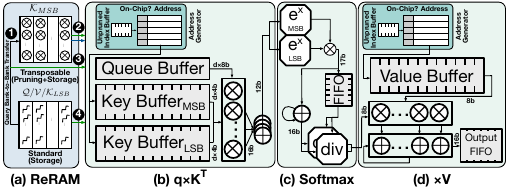}
\caption{The overview of \sys system. Data transfers within ReRAMs and between ReRAMs and accelerator consist of: (1) bank-to-bank between standard and transposable ReRAMs, (2) unpruned $\mathcal{K}_{MSB}$ vectors and their corresponding indices, (3) $\mathcal{Q}$ vector, and (4) unpruned $\mathcal{K}_{LSB}$ and $\mathcal{V}$ vectors.} %
\label{fig:overview} 
\end{figure}
To process the query vector $q_{1\times{\mathcal{S}}}$, the on-chip accelerator first transmits a subset of query vector MSBs to the transposable ReRAMs\footnote{The number of MSBs in query and key are identically set to 4-bits.} that store the key matrix $\mathcal{K}^T_{d\times{\mathcal{S}}}$.
A low-precision DAC converts the digital values of the query vector to analog and feeds them into the ReRAM via wordlines.
The transposable ReRAM array performs a low-precision vector-matrix multiplication in analog to calculate the \code{Scores}, which are produced after vertically applying an analog reduction sum per key vector.
The next step performs in-memory thresholding using analog comparators.
Note that the threshold values can either be set at the start of the computations or sent along with each $q$ vector.
Finally, after performing the analog comparisons, a voltage value (corresponding to a 1-bit digital value) flows through a series of 1-bit ADCs.
The ADC outputs indicate the pruning state of their corresponding key vectors, where ``1'' means pruned.

The generated binary pruning vector is sent back to the on-chip accelerator, which subsequently gets translated into multiple memory requests to selectively fetch unpruned key and value vectors from their corresponding modules.
Note that the pruning vectors for both key and value are completely identical. 
Upon receiving the first unpruned key vector, the accelerator can start recomputing \code{Score}s in full-precision.
The same process repeats for the rest of the query vectors.
\section{Overview of \sys System}
\label{sec:memPrune}
As shown in Figure~\ref{fig:overview}, the overall \sys architecture includes two main components: 1) ReRAM memory, and 2) on-chip accelerator, as described below.

\niparagraph{ReRAM memory.}
ReRAM memory banks are split into two categories, standard and transposable.
Standard ReRAM is solely used for storage ($\mathcal{Q}$, $\mathcal{V}$, and $\mathcal{K}_{LSB}$), while transposable ReRAM is for both storage ($\mathcal{K}_{MSB}$) and performing dot-product and in-memory thresholding, informing the on-chip accelerator which embedding vector to fetch.
The in-memory thresholding mechanism, explained in Section~\ref{subsec:inmemBkg}, exclusively uses MSBs to determine pruning criteria.

\niparagraph{On-chip accelerator.} 
\rev{The \sys on-chip accelerator performs three main operations, $q\times\mathcal{K}^T$, Softmax, and $\times\mathcal{V}$ in a pipelined manner. Figure~\ref{fig:overview} depicts the major microarchitectural units with their associated bit precision and data flow. The arithmetic is further described in Section~\Ref{sec:hardware_arch}.}
\rev{The accelerator fetches the unpruned $k$ / $v$ vectors from ReRAM along with a binary vector, indicating which $k$ / $v$ indices are unpruned, to store in unpruned index buffers.}
\rev{Based on these indices, the address generator block produces addresses to access the unpruned vector from $\mathcal{K}$/$\mathcal{V}$ buffers.
}
\rev{The accelerator first performs $q_{1\times{d}}\times\{\mathcal{K}_{MSB},  \mathcal{K}_{LSB}\}$ followed by an adder tree to precisely calculate the score values.
Note that, $\mathcal{Q}$ buffer only stores the streamed-in $q_{1\times{d}}$ temporarily for the window of score computation.}
\rev{Then, the Softmax block normalizes the Score values into a probability distribution proportional to the exponentials of the input Scores.}
\rev{Finally, the last block multiplies each $v$ vector by their corresponding Score probability, followed by a reduction sum across the weighted $v$ vectors to generate the final attention values.}

\section{\sys Memory Controller}
\label{sec:memCtrl}
\niparagraph{Background.}
\label{subsec:memBackground}
A memory controller receives a stream of memory access requests from the on-chip accelerator, generates their corresponding memory command stream.
The memory controller consequently arbitrates the memory commands and schedules them to off-chip memory according to a scheduling policy.
The technology of a memory (e.g. DRAM or ReRAM) dictates a set of timing constraints that must be satisfied by the memory controller between each issued memory command.
To communicate data between the on-chip accelerator and off-chip memory, a sequence of memory commands generated by the memory controller are required.
These commands collectively retrieve data from rows across multiple chips into their corresponding row buffers and select a column from the currently fetched retrieved data.
A subsequent column access to the same row enjoys the row-buffer locality, hence, lowest access latency. However, the consecutive accesses between different rows are generally suffer from substantially higher access latency.
The memory controller aims to schedule the memory commands in order to maximize the row-buffer locality.

\subsection{Data Layout Organization}
\label{subsec:dataLayout}
We presume a similar organization as conventional memory subsystems for \sys.
In general, optimizing the data layout organization for deep learning applications is straightforward because of their predictable memory access pattern. We observe the same pattern for \sys data layout organization.
As explained, to support in-memory thresholding, we presume a non-interleaving data organization for $\mathcal{K}$s (similar to prior work~\cite{gradpim:hpca:2021,farmahini2015nda,bufcomp:date:2016}).
That is, we store each vector of $k$ (a column in $\mathcal{K}^T_{d\times{\mathcal{S}}}$) in one column of memory mat.
Based on our observation (spatial locality between unpruned key indices, Section~\ref{subsec:motiv}), we distribute the neighboring $k$ vectors across different banks/channels.
Our empirical results show that this distribution of $k$ vectors provides a better utilization of memory bandwidth and reduces structural conflicts. Same data layout organization works for $v$ vectors.
The $\mathcal{Q}$ matrix, on the other hand, does not need to follow this particular data layout organization.
That is because each $q$ vector is processed sequentially and after every $q$-$\mathcal{K}$ vector-matrix multiplication which provides sufficient time for the memory subsystem to handle the upcoming query read requests.

The final data layout organization requirement is for the MSB and LSB parts of $k$ vectors.
As described in Section~\ref{subsec:transthr}, MSB and LSB parts of key vectors must be distributed across different type of ReRAM crossbar arrays, transposable and conventional respectively.
This separation of MSB and LSB bits can be established statically before the computation starts.
To effectively enable this special data layout organization, we can provide device-side allocation APIs so the user can specify different requirements for $\mathcal{Q}$ / $\mathcal{K}$ / $\mathcal{V}$ matrices without exposing physical underlying structure of memory subsystem. Similar software support has been proposed in prior work~\cite{gradpim:hpca:2021,choi2016multi}.

\niparagraph{Scaling for \rev{embedding size}.}
One potential challenge to the proposed data layout organization and in-memory thresholding mechanism is posed by scalability. 
Specifically, as the embedding size of key vectors increases, applying the reduction sum across each column of ReRAM arrays may seem infeasible.  
This limitation can be readily addressed by splitting the key vector into multiple adjacent ReRAM columns, similarly to ~\cite{kang2018multi}.
With this circuit modification, the resulting analog current from the adjacent key vector splits can be subsequently merged and compared with the threshold value.
\subsection{Memory Controller Microarchitecture}
\label{subsec:memCtrlOverview}
The on-chip memory controller designed for \sys is separated into a frontend and a backend engine.
The frontend engine communicates with multiple on-chip accelerators, accepting memory requests, whereas the backend engine generates and issues commands to off-chip memory modules with respect to their timing constraints.
\subsection{Memory Controller Execution Flow}
\label{subsec:memCtrlExec}
\niparagraph{Overview.}
The memory controller in \sys governs the tasks of in-memory thresholding and fetching the corresponding unpruned $d\times{1}$ vectors of $\mathcal{K}^{T}_{d\times{s}}$ matrix.
To complete these operations, the memory controller first sends a low-precision variant of $q_i$ vector of size $1\times{d}$ to $\mathcal{K}_{MSB}$ ReRAM banks.
Each $\mathcal{K}_{MSB}$ ReRAM bank executes low-precision in-memory thresholding and generates a binary pruning vector of size $s$.
The $j^{th}$ element of the generated binary vector indicates whether to prune the $j^{th}$ column of $\mathcal{K}^{T}_{d\times{s}}$ matrix (\emph{i.e.}, `1' $\rightarrow$ pruned and `0' $\rightarrow$ unpruned).
Upon receiving the binary pruning vector, the memory controller processes this vector and consequently issues a stream of read requests to fetch the unpruned vectors of $\mathcal{K}^{T}_{d\times{s}}$ matrix.

\niparagraph{Spatial locality detection engine.}
To further reduce the data movement between off-chip memory and on-chip buffers, we design and integrate a spatial locality detection (SLD) engine in the front-end of the memory controller.
The primary task of the engine is to detect and exploit spatial locality between the last and current binary pruning vectors associated with the attention score computations for adjacent query vectors (\emph{i.e.}, $q^i_{1\times{d}}$ and $q^{i+1}_{1\times{d}}$).
The advantages are two folds: (1) ``only'' generating memory requests for $k$ vectors that do not exists in on-chip $\mathcal{K}$ buffer, hence reducing data transfer and memory contention, and (2) bootstrapping the attention score ($\mathcal{Q}\times{\mathcal{K}^T}$) computations for the $k$ vectors that already reside in on-chip $\mathcal{K}$ buffer, hence minimizing the data transfer latency.
The following equations describe the logic behind these two tasks given the last and current binary pruning vector:
\begin{align}
\label{eq:score}
\text{\textbf{Task 1}$\rightarrow$ Memory Requests Vector} &= \mathcal{P}^{t-1}_{1\times{s}} \land \overline{\mathcal{P}^{t}_{1\times{s}}}\\
\text{\textbf{Task 2}$\rightarrow$ Spatial Locality Vector} &= \overline{\mathcal{P}^{t-1}_{1\times{s}}} \land \overline{\mathcal{P}^{t}_{1\times{s}}}
\end{align}
\noindent{}where $\mathcal{P}^{t-1}_{1\times{s}}$ and $\mathcal{P}^{t}_{1\times{s}}$ represent binary pruning vectors associated with the last and current attention score computations at a given time point $t$, respectively.

\niparagraph{Memory request generator engine.}
The main objective for the memory request generator (MRG) engine is to produce a potentially limited number of memory requests to fetch key vectors that do not currently reside in on-chip key memory.
Each memory controller retains one MRG engine to produce the corresponding key vector addresses residing in that particular bank.
At each cycle, a binary value is read from the memory request vector.
If zero, it means that the corresponding key vector is not required for the current attention score computation; hence, bypassing memory request generation step. 
On the other hand, a one-value indicates that a key vector must be fetched from off-chip memory. Hence, a memory request with an address corresponding to the location of the desired key vector is generated.

To satisfy the key vector organization requirement (Section~\ref{sec:memPrune}), we decided to statically place the adjacent key vectors into memory modules attached to different channels.
As such, to properly generate the key vector addresses, we equip each MRG with a base register and a shared up counter block.
The base register indicates the starting key vector index located on a particular memory channel.
The up counter starts from zero upon receiving a binary pruning vector and increases by the number of memory channels.
We also equip each memory controller with a key index generator (KIG) engine, which has the exact same microarchitecture.
However, in lieu of memory request vector, KIG engine operates on spatial locality vectors to generate the key vector addresses for \sys on-chip engines in order to bootstrap the attention score computations.

\niparagraph{Memory commands and timing considerations.}
Supporting \sys style in-memory thresholding into memory requires introducing additional memory commands and memory timing constraints.
To enable in-memory thresholding in \sys, we introduce two additional memory commands, \code{CopyQ} and \code{ReadP}.
\code{CopyQ} copies elements of query vector to in-memory query buffer, whereas \code{ReadP} reads elements of resulting binary pruning vector from in-memory pruning vector buffer.
Depending on the bit-width of query and pruning vectors, the memory controller may issue one or more consecutive \code{CopyQ} and \code{ReadP} commands.
Note that to initiate in-memory thresholding computations, we add one-bit in \code{CopyQ} command in which a one-value indicating the start of computations.
Issuing other memory commands will be prohibited amid in-memory thresholding computations.

As you may observe, there is some similarities between \code{CopyQ} and \code{ReadP} commands and normal memory read and write, respectively, projecting a similar timing constraints as read/write commands.
However, since \code{CopyQ} works with an isolated buffer from memory arrays, it neither requires \bench{tRP} for row pre-charging, nor \bench{tRCD} to activate a memory row.
On the other hand, since consecutive \code{CopyQ} commands still occupy data buses, we adhere to the \bench{tCL} timing constraint.
The scenario for \code{ReadP} is quite different as it communicates with the bank row buffers to read the resulting binary pruning vectors into on-chip buffers for further processing.
Therefore, we conservatively follow the exact same timing constraints as memory read command for \code{ReadP}.
For both introduced commands, burst \code{CopyQ} and \code{ReadP} follow the same timing constraints as normal burst memory read and write.

While the described scenarios for \code{CopyQ} and \code{ReadP} covers most of the required timing constraints, it still leaves one crucial timing constraints between adjacent \code{CopyQ} and \code{ReadP} commands.
This timing, dubbed \code{tAxTh}, represents the number of cycles that each ReRAM crossbar requires to perform in-memory thresholding and producing the resulting pruning vector.
Our circuit simulations show that this timing is <8 cycles \cite{naturememristor:2019}.

\niparagraph{Power implications of in-memory thresholding.}
In addition to timing constraints, memory systems are also under power budget limitations.
\code{tFAW} and \code{tRRD} represent the memory timing constraints linked to power budget.
To account for this power budget limitation, we model the analog in-memory thresholding circuit and estimate the power of analog comparators.
Our simulation shows that the overhead of additional analog circuitry for analog comparisons merely increases the total power budget by $<\,0.07\%$ of total in-memory computation~\cite{timely:2020}.
This power overhead has negligible implications on these timing constraints, hence we posit the nominal values for \code{tFAW} and \code{tRRD} in our simulations (similar to work~\cite{gradpim:hpca:2021,mvid:tc:2020}).
\section{\sys On-Chip Accelerator}
\label{sec:hardware_arch}
\begin{figure}
\centering
\includegraphics[width=0.99\columnwidth]{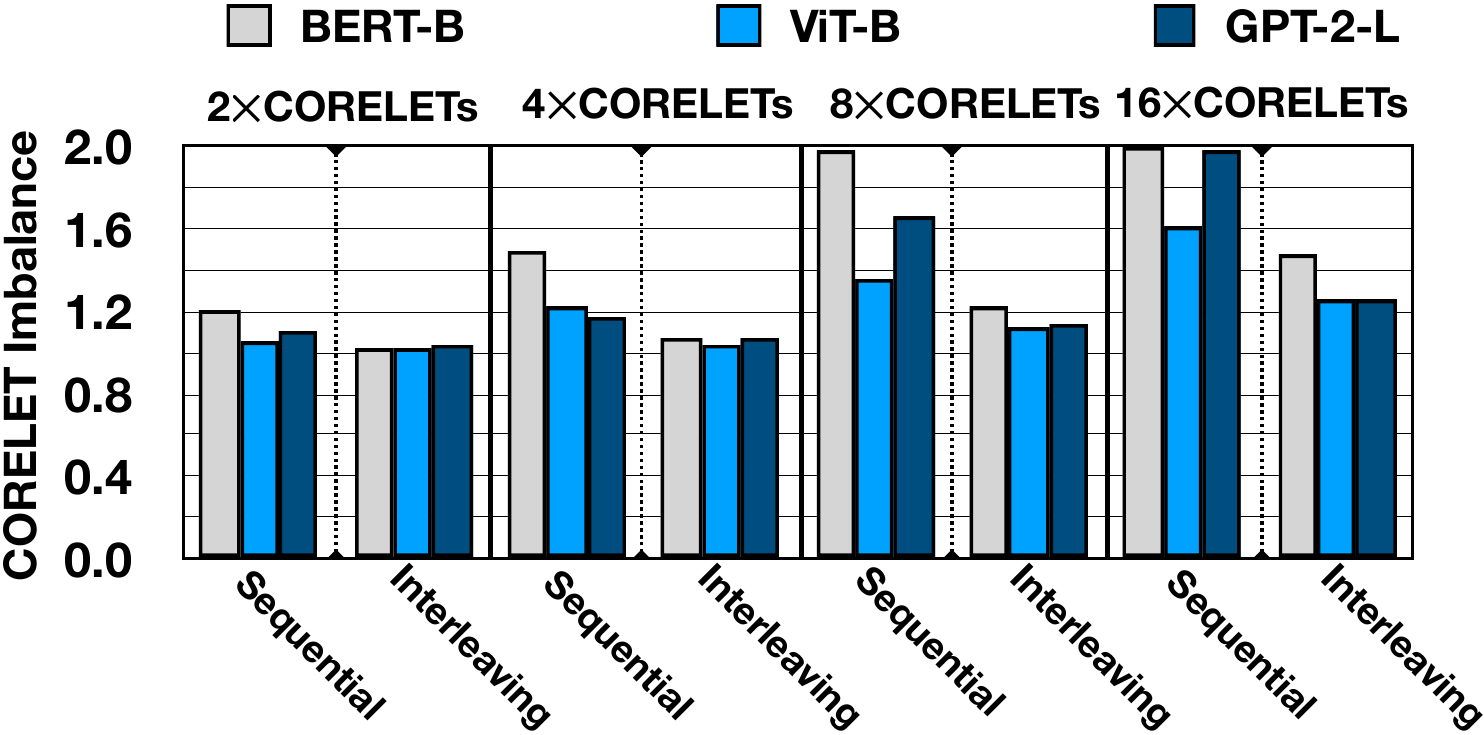}
\caption{\code{CORELET} utilization imbalance with and without token interleaving across \code{CORELET}s.}
\label{fig:utilization} 
\end{figure}
The \sys processor includes $N$ \code{CORELET}s to enable a higher parallelism degree.
\rev{A \code{CORELET} is an independent processing block that computes the entire self-attention mechanism pipeline, including $\mathcal{Q}\times\mathcal{K}^T$, Softmax, and $\times\mathcal{V}$.}
Each \code{CORELET} consists of a \code{QK}-processing unit (\code{QK-PU}) and a \code{V}-processing unit (\code{V-PU}).
\code{QK-PU} performs the $1 \times d$ dot product between $q$ and $k$, whereas \code{V-PU} processes the $1 \times d$  dot product between the Softmax output and $v$ in the digital domain.
In addition, each \code{CORELET} has a small number of buffers to store unpruned key and value vectors.
Note that the query vectors are processed in a stream manner, and thus do not need multi-entry buffers (\code{Q-buf}).
Finally, each \code{CORELET} has its own look-up-tables to record which key and value vectors are currently present on chip.

\niparagraph{Workload balancing across \code{CORELET}s.} 
\sys accelerator can simultaneously process multiple key vector sub-elements in each \code{CORELET} while the same query vector is distributed among all the \code{CORELET}s.
As soon as the computations of one query and all of its associated keys complete, the computations of the next query can begin.
In this design, the adjacent key vectors are assigned to different \code{CORELET}s, called \textit{token-interleaving}.
For example, given total four available \code{CORELET}s, \sys process $\mathcal{K}_{4n+i}$s in the $i$-th \code{CORELET} if the token is unpruned. 
This balances the workload across \code{CORELET}s while considering the spatial locality, by which the unpruned indices tend to appear in adjacent locations.

Figure~\ref{fig:utilization} shows the workload imbalance ratio \rev{with 2$\times$, 4$\times$, 8$\times$, and 16$\times$ \code{CORELET}s}.
We calculate the imbalance ratio by dividing the maximum by the minimum numbers of assigned unpruned tokens per \code{CORELET} and averaging the numbers for all the queries (i.e. the value of one implies ideal workload balance across \code{CORELET}s). 
The proposed workload distribution scheme considerably improves the utilization balance compared to the sequential token mapping, e.g. neighboring tokens belonging to the same \code{CORELET}s. 
\rev{We observe that for large models such as \bench{GPT-2-L}, \sys can readily leverage higher parallelism from more \code{CORELET}s. However, \sys may underutilize the \code{CORELET}s for smaller models (e.g., \bench{BERT-B}) because of their inferior parallelism opportunity}.

\niparagraph{Handling data misses.}
To minimize the number of stalls due to data misses, the unpruned key vectors are proactively prefetched by the memory controller (as explained in Section~\ref{subsec:memCtrlOverview}).
We also configure the main memory bandwidth (Table \ref{table:arch_config}) to provide a new pair of $k$ and $v$ in burst mode to further reduce such stalls.
Note that by leveraging the spatial locality between unpruned key vectors, on average, only 2.1\% of the sequence length is required to be fetched between adjacent queries.
This high data reuse drastically reduces the likelihood of data misses.
When a rare data miss occurs, the computations for the next available key vector can proceed until the data miss is handled by the memory controller. 
We implement this bypassing of unavailable key vectors by adding a rotating pointer to key/value index buffers.

\niparagraph{\sys accelerator arithmetic operations.}
Once at least one key vector resides in \code{K-buf}, the computation can start.
At each cycle, \sys performs a dot-product between each subset of elements from key and query vectors.
If all the key elements can not be processed during one compute iteration, \sys stores the partial sums in a register until the results are ready to be processed by a Softmax module.
Similar to prior work~\cite{a3:hpca20, leopard:isca:2022}, we use a two look-up-tables method for exponent calculation.
Afterwards, \sys stores the streaming outputs in FIFOs for accumulation.
Once complete, each score is normalized to produce the corresponding probabilities.
To balance the throughput between different stages of the pipeline, we employ two divider units.
Finally, the computed score probabilities are used in \code{V-PU} to calculate the weighted sum of $v$ vectors.
Note that the unpruned indices for key vectors can be used for the pruning of value vectors as well.

\niparagraph{Two-dimensional sequence reduction.}
As introduced in Section~\ref{sec:padding}, a large portion (e.g. 46\% for the \bench{SQUAD} dataset) of the total sequence length is futile due to zero-padding.
Figure \ref{fig:spatial locality}  illustrates the zero-padded (gray) area, which reduces the required output computation in both vertical and horizontal dimensions.
Horizontally, the computation is reduced to $k$ vectors per $q$, whereas vertically, it is reduced to $q$ vectors. 
We implement this mechanism by enabling the memory controller to filter out the read requests for these masked regions.

\niparagraph{\sys accelerator design choice.}
The \sys accelerator does not employ a double-buffering scheme for on-chip memory in order to avoid the doubled cost of memory capacity.
When the new data arrives from main memory, those are stored in a temporary small buffer. Meanwhile, a stall request is issued to initiate the write process into \code{K-buf} and \code{V-buf}.
Note that, due to spatial locality across unpruned $k$ elements for adjacent $q$ vectors, the number of newly fetched $k$ / $v$ is infrequent.
Similar to prior work~\cite{leopard:isca:2022,spatten:hpca21,a3:hpca20}, \sys performs all the computations in 8-bit precision, except Softmax with 12-bit inputs. For final attention score, we employ 16-bit precision.
%
\section{Methodology and Evaluation}
\label{sec:eval}

\niparagraph{Benchmarks.}
We use the following models to evaluate the efficacy of \sys: \bench{BERT-Base} (\bench{BERT-B})~\cite{gbert:naacl:2019},  \bench{BERT-Large} (\bench{BERT-L})~\cite{gbert:naacl:2019},
\bench{ALBERT-X-Large} (\bench{ALBERT-XL}) \cite{albert:iclr:2019}, 
\bench{ALBERT-XX-Large} (\bench{ALBERT-XXL}) \cite{albert:iclr:2019},
\bench{ViT-Base} (\bench{ViT-B}) \cite{chen2021chasing},
and \bench{GPT-2-Large} (\bench{GPT-2-L}) \cite{gpt2}.
We employ the Stanford Question Answering Dataset (\bench{SQUAD}) \cite{squad:arxiv:2016} to test \bench{BERT-B}, \bench{BERT-L}, \bench{ALBERT-XL}, \bench{ALBERT-XXL}, the \bench{WikiText-2}\cite{wikitext2} to test \bench{GPT-2-L}, and \bench{CIFAR10} \cite{cifar10} dataset to test the \bench{ViT-B}.
We use the default sequence length ($s$) of 197, 384, and 1024 for the \bench{CIFAR10}, \bench{SQUAD} and \bench{WikiText-2} datasets,   respectively.
All models use an embedding size of $d=64$.
On top of the above datasets, we create two additional synthetic models \bench{Synth1} and \bench{Synth2} with 2K and 4K sequences.
These additional models estimate the projected benefit of \sys architecture for longer input sequences.

\niparagraph{Model fine-tuning for target benchmarks.} 
For the baseline, we use pre-trained models from HuggingFace~\cite{huggingface:2019} and fine-tune them on each task with the reported hyperparameters~\cite{gbert:naacl:2019,albert:iclr:2019,vit}.
We only alter batch size due to our limited GPU memory.
Nonetheless, even with reduced batch size, the final accuracy after fine-tuning does not change discernibly.
Following the described methodology~\cite{leopard:isca:2022}, we implement differentiable soft thresholding into the studied transformer models.
We identify the optimal pruning threshold per attention layer as part of the task-specific fine-tuning process.
We use identical hyperparameters for training, except for the learning rate and the number of epochs.
The search space for the model learning rate is \{$2e^{-3}$, $2e^{-4}$, $2e^{-5}$\}, whereas the search space of \{$2e^{-5}$, $2e^{-6}$\} is used for the learned threshold.
The number of epochs varies from one to three depending on the target tasks. 
We conduct our experiments with PyTorch v1.10~\cite{pytorch} on an Nvidia RTX 3090 GPU, except for \bench{GPT-2-L}, for whcih we use an Nvidia A100 GPU.

The resulting pruning rates for \bench{BERT-B}, \bench{BERT-L}, \bench{ALBERT-XL}, \bench{ALBERT-XXL}, \bench{ViT-B}, and \bench{GPT-2-L} are 74.6\%, 75.5\%, 65.1\%, 73.1\%, 64.4\%, and 73.9\%, respectively.
For \bench{Synth1} and \bench{Synth2}, we set a pruning rate of 75\% and a padding ratio of 50\%.
The estimated main memory access when switching to new query vector for \bench{Synth1} and \bench{Synth2} are obtained by scaling up the numbers from \bench{BERT-B} based on the sequence length difference.
\begin{table}[t!]
\footnotesize
\centering
\vspace*{0.1cm}
\caption{\label{table:arch_config}Hardware configurations of \sys.}
\resizebox{0.48\textwidth}{!}{
\begin{tabular}{l|l}
\toprule

\textbf{Modules} & \textbf{Configurations for S-\sys~/  M-\sys~/ L-\sys} \\\bottomrule
\rev{ReRAM} BW &  $16 \times 64$-bit channels @ 1 GHz per \code{CORELET} \\\hline
ReRAM Array & 256 $\times$ 128 standard bitcell, 64 $\times$ 128 transposable array with 4-b MLC\\\hline
On-Chip Cache & 16/32/64KB in total \rev{of K/V buffers} (= 8/16/32 banks), 128-b port per bank \\\hline
QK-PU / V-PU & 1/2/4 EA of 1-D 64 (=$\mathcal{D}$) way 8$\times$8-b MAC array\\\hline
Softmax & 1/2/4 EA of 12-b input,  8-b output, 2EA of 64B LUTs, 2EA of dividers\\\hline
\rev{Query Buffer}&\rev{64B / 128B / 256B}\\\hline
\rev{Index Buffer}&\rev{0.5KB / 1KB / 2KB}\\\bottomrule
\end{tabular}
}
\end{table}

\niparagraph{\sys hardware simulations.}
\label{para:nvsim}
Table~\ref{table:arch_config} lists the design parameters of \sys for three studied configurations:
(1) S-\sys: a \code{CORELET} with 16KB, 
(2) M-\sys : two \code{CORELET}s with 32KB, 
and (3) L-\sys: four \code{CORELET}s with 64KB total on-chip buffer capacity.
We use Cadence Genus 19.1~\cite{genus} for the logic synthesis and Cadence Innovus 19.1~\cite{innovus} for the placement/routing (PnR) of digital blocks with a 65~nm TSMC general-purpose standard cell library.
We generate the digital blocks to meet the target frequency of 1\,GHz from the post-layout simulations.
For SRAM on-chip memories, we use ARM Memory Compiler with High density 65\,nm single-port SRAM (version r0p0) \cite{mem_compiler} to measure its energy consumption.
ReRAM crossbar in-memory operation consumes 0.10 pJ\,/\,MAC in 65\,nm including the digital-to-analog conversion (DAC)~\cite{naturememristor:2019}.
The standard ReRAM read/write operations consume 3.1\,pJ/bit and 24.4\,pJ/bit, respectively~\cite{rwdevice2019}\footnote{\rev{Compared to NVSim~\cite{nvsim:tc:2012} with a similar configuration, we use a more conservative model with 1.6$\times$ and 7.2$\times$ higher read delay and energy, respectively, to accommodate for additional ReRAM overheads.}}.
Each analog comparator consumes 41\,fJ~\cite{timely:2020}.
A recent study of ReRAM in-memory computing~\cite{1t1mmultiply:2016} has shown that 64-tap in-memory dot-product delivers 5-bit equivalent output accuracy.
To emulate the limited accuracy of the in-memory thresholding, we use an identical error specification in Section~\ref{subsec:inmemth} with $b=5$.

\begin{table}[t!]
\footnotesize
\centering
\vspace*{0.1cm}
\caption{\label{table:perf_sim}Energy consumption of major microarchitectural units of \sys.}
\resizebox{0.48\textwidth}{!}{
\begin{tabular}{l|l}
\toprule
\textbf{Microarchitecture Units} & \textbf{Energy} \\\bottomrule
QK-PU/V-PU Dot-Product&192.56 (pJ); 8 bits, 64-tap\\\hline
Key/Value Buffer&256 (pJ); 4 banks with 128-bit access per bank\\\hline
Softmax&89.8 (pJ); 2 LUT accesses + multiply + division\\\hline
Analog Comparator&5.34 (pJ); 128 Columns\\\hline
In-Memory Computation&833.6 (pJ); 64 Rows$\times$128 Columns\\\hline
ReRAM Access&Write: 12492.8 (pJ), Read: 1587.2 (pJ); 512 bits\\\bottomrule
\end{tabular}
}
\end{table}

\niparagraph{\rev{\sys performance simulator.}}
\label{para:perfsim}
We collect the numbers of in-memory dot-product operations and analog comparisons, read accesses to ReRAM, on-chip $q\times\mathcal{K}^T$ for unpruned elements, accesses to LUTs, and division operations for Softmax.
In addition, we compile the numbers of additions and multiplications to calculate the weighted sum of $v$ vectors.
For the numbers of read accesses to ReRAM, the simulator properly accounts for the spatial locality between adjacent queries and limited on-chip memory capacity.
Finally, because the majority of these statistics are input-dependent, we average these numbers across the entire input dataset for each model.
For energy consumption, we multiply the average number of operations across the self-attention layers by their corresponding energy consumption from post-layout simulation along with the reported access energy for ReRAM, as listed in Table~\ref{table:perf_sim}. 
For latency and throughout estimation, we calculate the delay of self-attention layers while taking into account the in-memory thresholding compute delay along with the communication delay between ReRAM arrays and \code{CORELET}s.
We also implement token interleaving (See Section~\ref{sec:hardware_arch}), which distributes tokens across the entire input dataset to \code{CORELET}s.
Such interleaving better leverages the spatial localities between $k$ vectors and minimizes the imbalance factor.
We report the delay of each self-attention layer as the worst-case delay across the $N$ \code{CORELET}s.

\niparagraph{Baseline architecture.}
We employ the same configuration of S-\sys, M-\sys, and L-\sys, but without the in-memory pruning, proposed memory controller, and two-dimensional computing reduction for the padded sequences.
We compare \sys and the baseline at iso-setups including the same frequency, the number of processing elements, on-chip memory capacity, and  bit widths for all the input and output of digital logic blocks.

\niparagraph{Comparison to prior systems.}
\aaa~\cite{a3:hpca20}, SpAtten~\cite{spatten:hpca21}, and LeOPArd \cite{leopard:isca:2022} also support the run-timing pruning to minimize the required computation.
\aaa is a prior state of the art on using approximation to accelerate attention mechanism.
\aaa thresholds after processing a limited number of $k$ vectors from the sorted queue in a magnitude order to minimize the run-timing pruning overhead.
Nonetheless, \aaa does not consider the data movement cost from the main memory assuming enough on-chip memory capacity. 
LeOPArd performs the gradient-based training to co-optimize the model accuracy and pruning rate by tuning the pruning threshold automatically during the training instead of empirical methods. Again, LeOPArd does not consider the cost from main memory access.
SpAtten proposed a cascaded pruning to exclude the redundant heads and tokens from all the subsequent layers once those are pruned in the previous layer.
SpAtten reduces the DRAM access cost for \bench{GPT-2}, but not for other models assuming enough on-chip capacity.
\begin{figure}
\centering
\includegraphics[width=0.99\columnwidth]{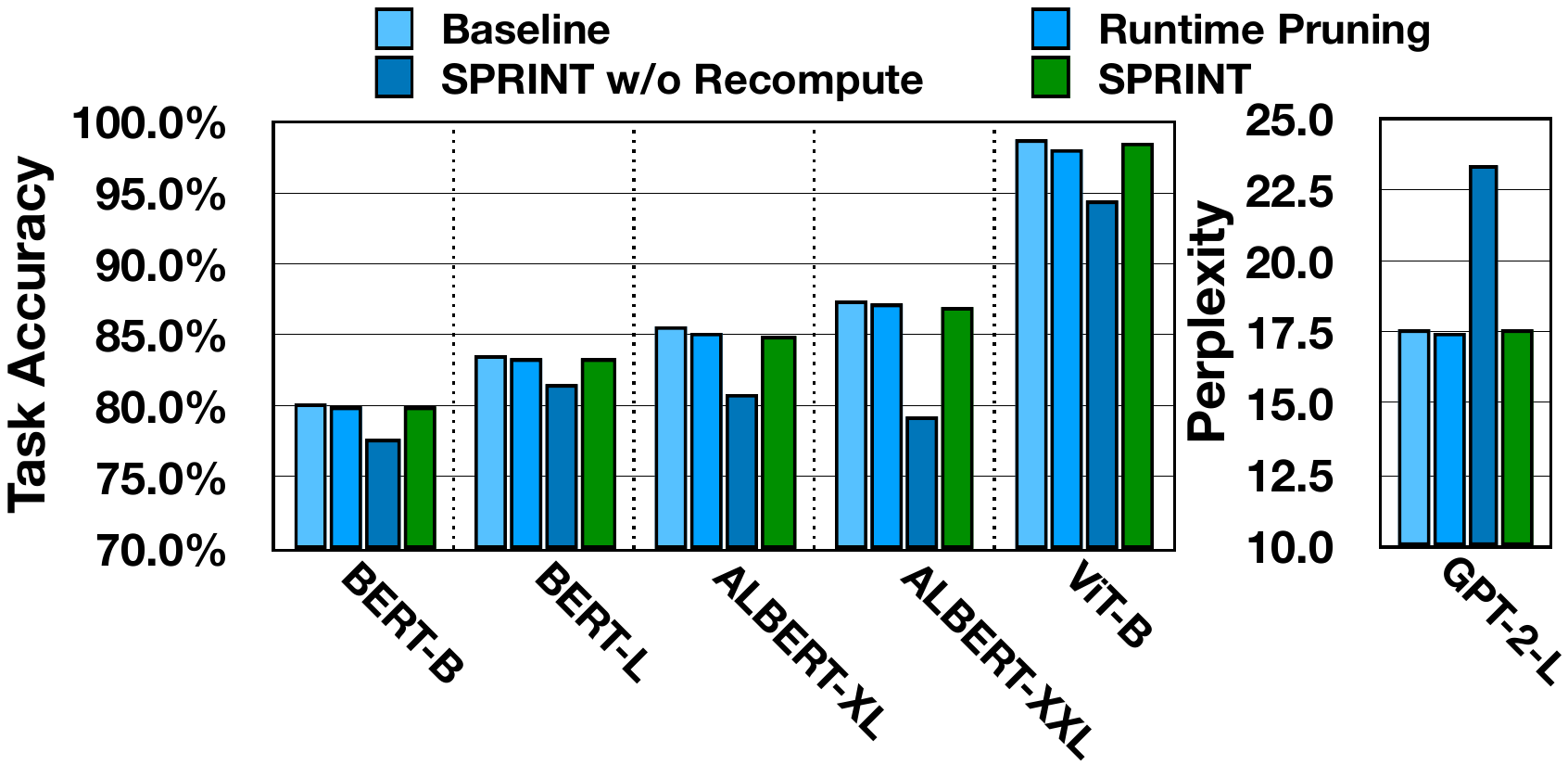}
\caption{\rev{Comparison of task accuracy between baseline and different \sys configurations. Second and third bars show task accuracy with runtime pruning and \sys w/o on-chip recompute. Fourth bar depicts \sys accuracy after including on-chip recompute. \bench{GPT-2-L} accuracy is measured as a perplexity metric (lower is better).}}
\label{fig: accuracy} 
\end{figure}
\begin{figure*}
\centering
\includegraphics[width=2.05\columnwidth]{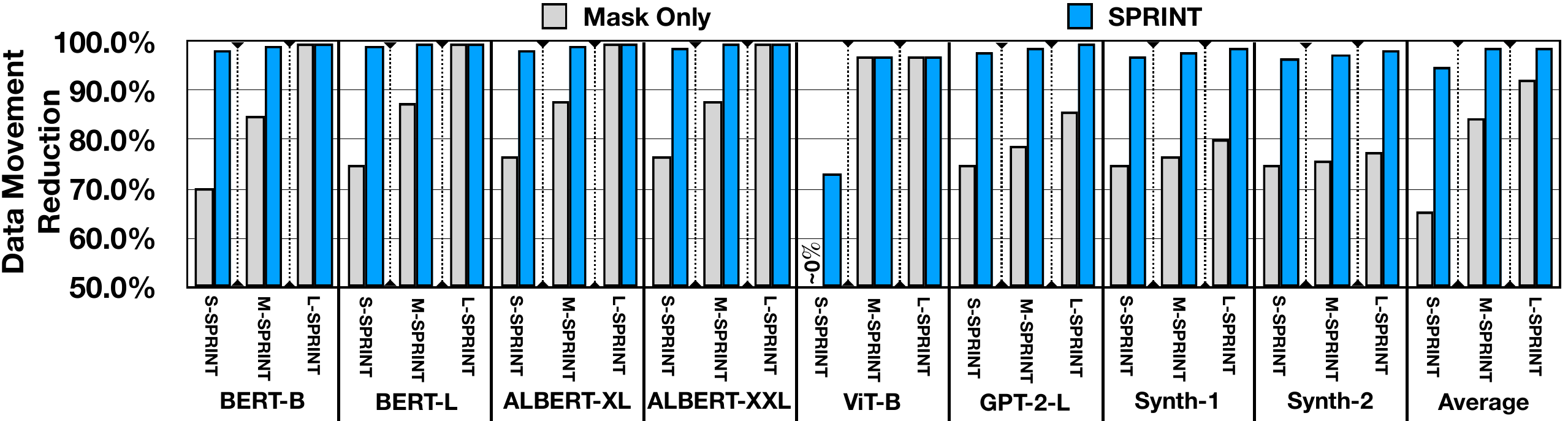}
\caption{Total data movement reduction from main memory normalized to that of S-Baseline configuration in two scenarios: (1) ``mask only'' $\mapsto$ with sequence reduction for the padded area and (2) ``\sys'' $\mapsto$ with run-time pruning on top of the sequence reduction.}
\label{fig:data_fetch} 
\end{figure*}

\subsection{Accuracy and Performance} 
\label{subsec:perf_results}
\niparagraph{Impacts on model accuracy from in-memory pruning.} 
Figure~\ref{fig: accuracy} depicts the model accuracy of the various models under four different scenarios: (1) baseline (software-only)~\cite{huggingface:2019}, (2) with runtime pruning, (3) \sys without on-chip recompute, and (4) \sys that includes both in-memory thresholding and on-chip recompute.
On average, the absolute accuracy difference (excluding \bench{GPT-2-L}) for runtime pruning (second bar) and \sys (fourth bar) is 0.22$\%$ with maximum degradation of 0.24$\%$ in \bench{ALBERT-XL}.
Compared to runtime pruning, \sys improves \bench{ViT-B} accuracy by 0.5$\%$.
Figure~\ref{fig: accuracy} also ablates the impact of on-chip recompute (third bar) on accuracy.
On average, the accuracy degradation of \sys without on-chip recompute is $\approx{}4\%$ ($5.71$ perplexity gap in \bench{GPT-2-L}).
On average, compared to the baseline models (first bar), accuracy degradation of \sys is merely \AverageAccuracyDegradation and separately, the perplexity of \bench{GPT-2-L} increases by 0.10.
These results underscore the importance of on-chip recompute in preserving the baseline model accuracy.

\niparagraph{Main memory data movement analysis.}
Figure \ref{fig:data_fetch} shows the reduction in the total amount of data movement from the main memory to the processor (compared to S-Baseline) during processing a single self-attention head.
We illustrate the data movement reduction in two configurations: (1) ``Mask Only'' $\mapsto$ sequence reduction for the padded area and (2) ``\sys'' $\mapsto$ run-timing pruning on top of the sequence reduction of the padded area. 
We normalize the results to S-Baseline, in which neither of these optimizations are employed.
The data reduction is higher with L-\sys due to the large on-chip buffer whereas  S-\sys requires more data movement.
Across the 48 studied configurations, our proposed system yields, on average, \DataReductS, \DataReductM, and \DataReductL~in S-\sys, M-\sys and L-\sys, respectively.
The benefit varies across workloads due to their different pruning rate and the portion of padded area.
For instance, \bench{BERT-B} has higher data movement reduction due to its 46\% padded area and 74.6\% pruning rate compared to \bench{ViT} with 64.3\% pruning rate and no padded area.
The mask only scenario has the modest data movement reduction of \MaskReductS, \MaskReductM, and \MaskReductL~in S-\sys, M-\sys, and L-\sys, respectively.
The only exception is observed in \bench{ViT-B} due to the lack of zero padding.
Specifically, \bench{ViT-B} is an exception because  M-\sys has already sufficient memory capacity to store entire (197) sequence length.
The gap among S-, M-, L-\sys configurations is the narrowest in \bench{Synth1} and \bench{Synth2}, where the input sequence length is significantly larger.
Thus, even L-\sys model can accommodate only highly limited fraction of the entire sequence length, e.g. 12.5\% in \bench{Synth2}.
For the same reason, the data movement reduction is less significant in \bench{Synth} models compared to others as those cannot contain enough number of correlated tokens in their scarce on-chip memory.
\begin{figure}
\centering
\includegraphics[width=0.99\columnwidth]{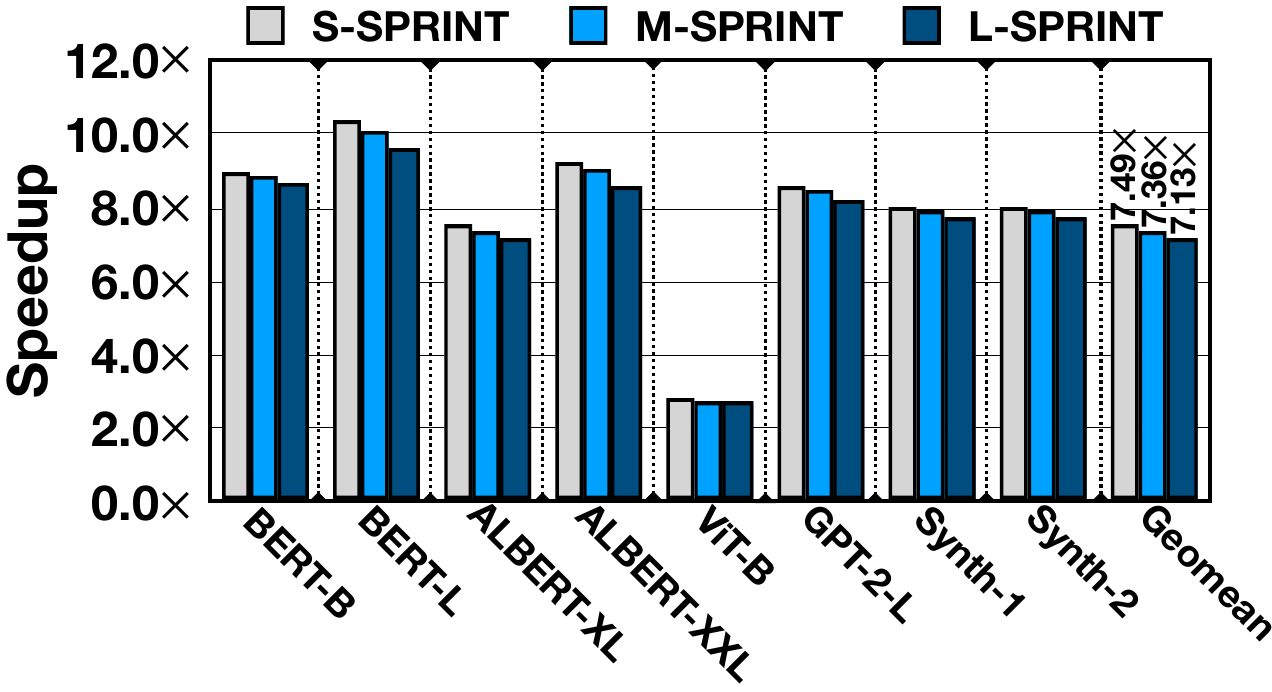}
\caption{Speedup comparison to a baseline design for \rev{self-attention layers}.}
\label{fig:speedup} 
\end{figure}
\begin{figure}
\centering
\includegraphics[width=0.99\columnwidth]{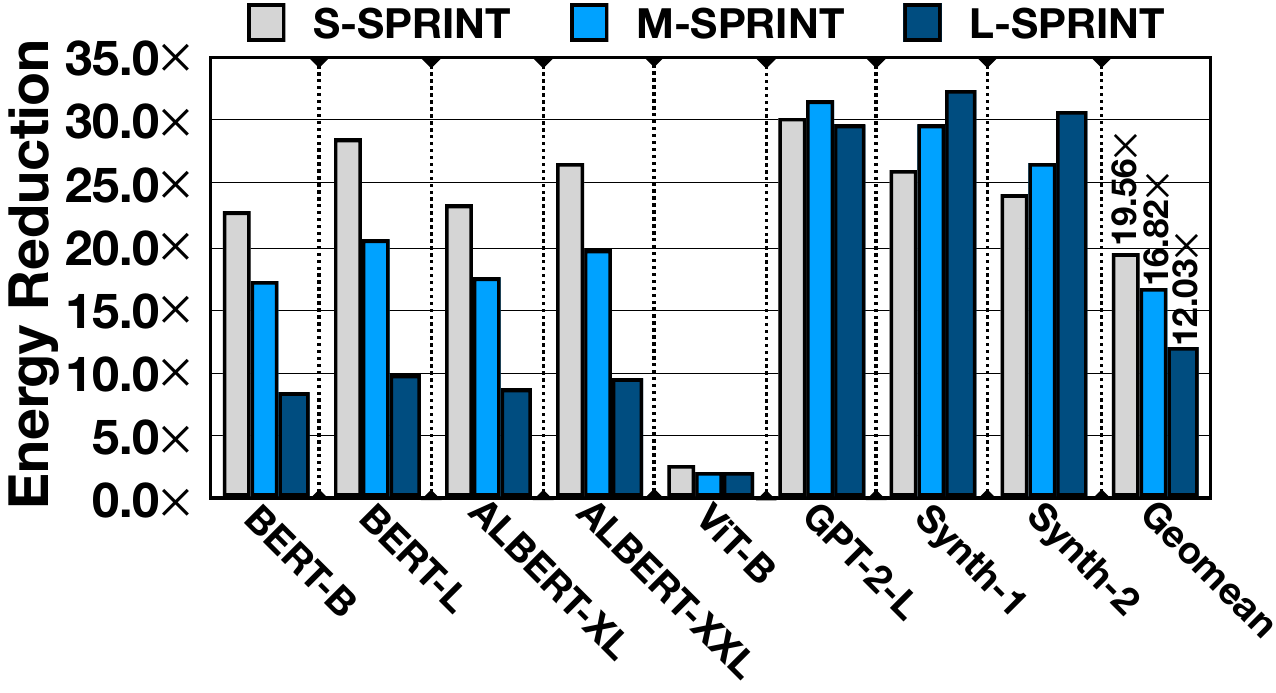}
\caption{Total energy reduction compared to a baseline design for \rev{self-attention layers}.}
\label{fig:energy} 
\end{figure}

\niparagraph{Performance and energy comparison.}
Figure~\ref{fig:speedup} compares the \sys speedup over the baseline design across all the 24 studied task. 
On average, S-, M-, and L-\sys achieve \SpeedupOverBaselineS, \SpeedupOverBaselineM, and \SpeedupOverBaselineL speedup, respectively.
These speedups are attributed to skipping the majority of computing cycles due to the in-memory run-time pruning. 
From the ablation study, the limited speedup of 1.8$\times$, 1.7$\times$, and 	1.7$\times$ is achieved on average from the run-time pruning without the in-memory computing support.
This is because all the $\mathcal{Q}\times\mathcal{K}^T$ must initially be computed in the on-chip accelerator so the $\times \mathcal{V}$ processing can be pruned.
The speedup benefit diminishes in L-\sys slightly (<5\% compared to S-\sys)  because the workload utilization is not appropriately balanced across \code{CORELET}s (See Figure~\ref{fig:utilization}) even after $k$ vector distribution.
\bench{BERT-L} enjoys the maximum benefits with 9.6 - 10.4$\times$ speedup while \bench{ViT-B} has minimum improvement of 2.7 - 2.8$\times$.
This is because of the different pruning rates and portion of padded area in those models. 

Figure~\ref{fig:energy} shows the energy reductions achieved by \sys, including  on-chip accelerator and ReRAM-based main memory, compared to the Baseline for the three configurations.
We observe an energy reductions  of \EnergyOverBaselineS ~for S-\sys, \EnergyOverBaselineM ~for M-\sys, and \EnergyOverBaselineL ~for L-\sys. 
The S-\sys achieves the largest energy reduction because the proportion of main memory access out of the total energy is significantly higher than the other configurations.
We attribute this higher main memory accesses to the highly constrained memory capacity and frequent memory accesses.
This leads to more improvement by the proposed technique which reduces the data movement effectively.    
On the other hand, \bench{Synth1} and \bench{Synth2} show the exception in this trend because even L-\sys can contain only very few fraction of the entire sequence, e.g. 12.5\% in \bench{Synth2} vs. 100\% in \bench{BERT-B}. 
In such a regime, where the memory capacity is significantly limited, the larger memory provides more room to fetch the correlated data together increasing the chance of data re-use. 
Therefore, L-\sys achieves more energy benefit compared to S- and  M-\sys in \bench{Synth} models.
The energy benefit is greater in \bench{Synth1} and \bench{Synth2} models than the other cases as those require more frequent data access due to their large sequence length so that the benefit by \sys is magnified. 
In contrast, \bench{ViT-B} shows the minimum benefit due to its small sequence length, and thus infrequent data access.
\begin{figure}[t]
\centering
\subfloat[Baseline vs. Pruning vs. \sys]{\label{fig:energy_saving_contribution_main}\includegraphics[width=0.99\columnwidth]{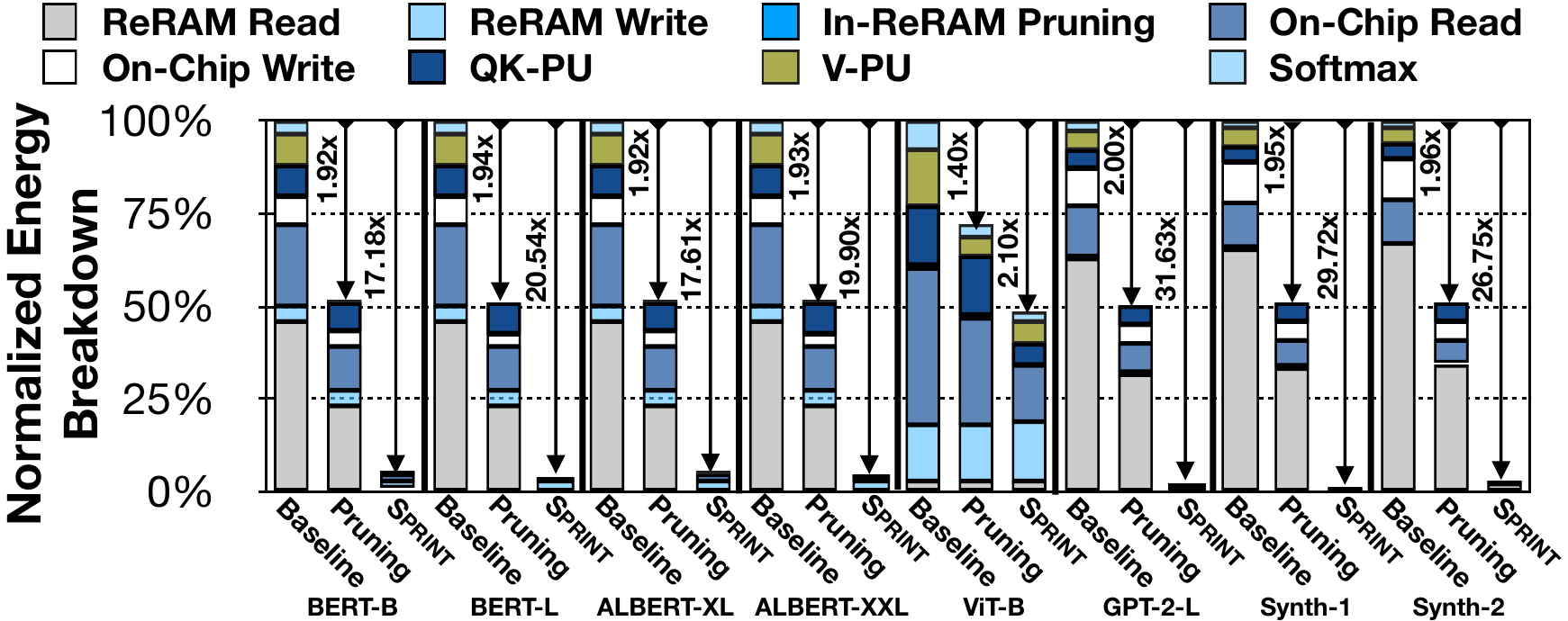}}\\
\subfloat[\sys Zoom-In]{\label{fig:energy_saving_contribution_zoom}\includegraphics[width=0.99\columnwidth]{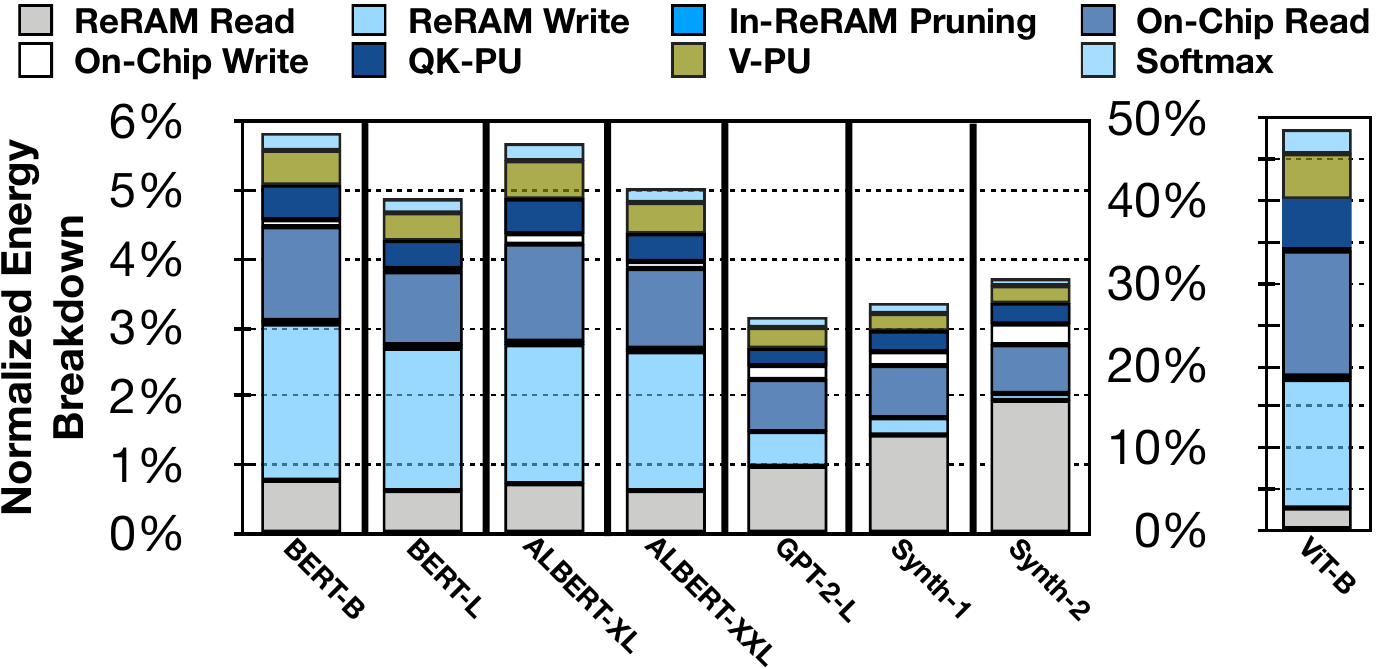}}
\caption{M-\sys's energy breakdown normalized to baseline \rev{for self-attention layers}. (a) The first bar shows the energy breakdown for baseline (no pruning), whereas the second and the third bars present the energy breakdown for pruning-only and \sys (in-ReRAM pruning), respectively. (b) Zoomed-in view of normalized energy breakdown for \sys.}
\label{fig:energy_saving_contribution} 
\end{figure}

\niparagraph{Energy consumption breakdown.}
Figure~\ref{fig:energy_saving_contribution_main} details the energy breakdown of M-\sys with pruning-only (second bar) and with pruning+in-ReRAM thresholding (third bar).
The energy breakdown includes: (1) ReRAM read/write, (2) in-ReRAM pruning, (3) on-chip $\mathcal{K}$/$\mathcal{V}$ buffers read/write, and (4) computations in \code{QK-PU}, Softmax, and \code{V-PU}.
On average, in baseline (first bar) 47.8$\%$ of the energy consumption comes from ReRAM read operations, except \bench{ViT-B}, whose input sequence is 2$\times$-~5$\times$ shorter compared to other models.

For the pruning-only scenario, \sys still needs to fetch the entire $q$ and $k$ vectors from ReRAM, even though some are inconsequential, and perform the requisite $\mathcal{Q}\times\mathcal{K}^T$ computations followed by on-chip comparison with threshold values for the run-time pruning.
After that,  Softmax and $\times \mathcal{V}$ are processed for the unpruned tokens.
Therefore, most benefits in the pruning-only case originate from reducing the number of main memory and on-chip memory reads for $\mathcal{V}$ and operations in Softmax and dot-product in \code{V-PU}.
Across the self-attention models, we observe around 1.9$\times$-~2.0$\times$ energy savings, except \code{ViT} with only 1.4$\times$.
We attribute the lower energy savings in \bench{ViT-B} to its lower pruning rate (64$\%$), fewer spatial localities (on average 2.6$\times$ less compared to other models), and lack of masking (See the gray stripes in Figure~\ref{fig:spatial locality}).

With pruning and in-ReRAM threshold (third bar), \sys significantly reduces the number of ReRAM reads as well as on-chip computations ($\mathcal{Q}\times\mathcal{K}^T$, Softmax, and $\times\mathcal{V}$), mainly by the virtue of in-ReRAM pruning.
In this configuration (zoomed-in view in Figure~\ref{fig:energy_saving_contribution_zoom}), \sys only fetches the elements of $\mathcal{K}$ and $\mathcal{V}$ matrices (unpruned ones) that certainly contribute to the computation of final attention values.
On average, \sys reduces the overall energy consumption of self-attention layers by 16.9$\times$.
Compared to other models, \bench{ViT-B} confers significantly lower energy savings, merely 2.10$\times$, for the same reasons as the pruning-only scenario.
While \sys architecture slightly reduces the number of ReRAM writes by obviating this need for zero-padded areas, ReRAM writes still have the highest contribution to the overall energy consumption.
The overhead of in-memory pruning including peripheral circuitry is negligible (only \ReRAMEnergyOverhead) due to its highly parallel and low-voltage analog operations.
These results substantiate that the in-ReRAM pruning benefits outweigh its marginal overhead.

\niparagraph{Data movement cost analysis.}
\label{para:datamove}
The data movement described in Figure \ref{fig:overview} is categorized (1)  bank-to-bank transfer of $\mathcal{Q}$ (\textcolor{black}{\ballnumber{1}} in Figure \ref{fig:overview}), and the transfer from the ReRAM main memory to the on-chip accelerator for (2)  $\mathcal{Q}$ \textcolor{black}{\ballnumber{3}}, and (3) unpruned  $\mathcal{K}$ \textcolor{black}{\ballnumber{2}}, \textcolor{black}{\ballnumber{4}} and $\mathcal{V}$ \textcolor{black}{\ballnumber{4}}.
We include the contributions of these data movements in the analysis of Figure \ref{fig:energy} and Figure \ref{fig:energy_saving_contribution}. 
The energy overhead of (1) is negligible ($<0.04$\% of the entire energy based on the post-layout simulations with an activity factor of 0.5) as it is the intra-chip data transfer. 
However, on average, (2) takes $<$3.7\% whereas (3) consumes 31.2\%, 20.7\%, and 15.5\% of energy for S-/M-/L-\sys, respectively.
\begin{table}
\footnotesize{
\centering
\vspace*{0.1cm}
\caption{\label{table:comparison}\sys performance comparison  with prior work. \sys and \textbf{LeOPArd} use the 65\,nm technology node,  whereas \textbf{A$^\mathrm{3}$} and \textbf{SpAtten} use the 40\,nm version.}
\resizebox{0.49\textwidth}{!}{
\begin{tabular}{l|c|c|c|c}
\bottomrule
\multirow{2}{*}{\textbf{Metric~(unit)}}&\multirow{2}{*}{\textbf{A$^\mathrm{3}$}}&\multirow{2}{*}{\textbf{SpAtten}}&\multirow{2}{*}{\textbf{LeOPArd}}&\multirow{2}{*}{\textbf{M-\sys}}\\
&&&&\\\bottomrule
\textbf{Sequence~Length}&50 - 384&384 - 1024&50 - 1024&128 - 4096\\\midrule
\textbf{Process~(nm)}&40&40&65&65\\\midrule
\textbf{Area~(mm$^2$)}&2.1&1.6&3.5&1.9\\\midrule
\textbf{Key Buffer~(KB)}&20&24&48&16\\\midrule
\textbf{Value Buffer~(KB)}&20&24&64&16\\\midrule
\textbf{GOPs~/~s}&518.0&360.0&574.1&1816.2\\\midrule
\textbf{GOPs~/~J}&4709.1&382.0&519.3&902.7\\\midrule
\textbf{GOPs~/~s~/~mm$^\mathrm{2}$}&249.0&238.0&165.5&973.5\\\midrule
\textbf{\rev{GOPs~/~s~/~J~/~mm$^\mathrm{2}$}}&\rev{2263.6}&\rev{252.5}&\rev{119.7}&\rev{469.7}\\\midrule
\textbf{Mem.~Cost~Included}&\normalsize{\color{red}{\XSolidBrush}}&\normalsize{\color{OliveGreen}{\CheckmarkBold}}&\normalsize{\color{red}{\XSolidBrush}}&\normalsize{\color{OliveGreen}{\CheckmarkBold}}\\\bottomrule
\end{tabular}
}
}
\end{table}

\niparagraph{Comparison with \aaa, SpAtten, and LeOPArd.}
Table~\ref{table:comparison} lists the details of prior works and M-\sys architecture in terms of throughput (GOPs~/~s), energy efficiency (GOPs~/~J), and area efficiency (GOPs~/~s~/~mm$^2$). For fair comparison, we also included the area from the in-memory thresholding \cite{trans:isscc:2020}, which takes only 3\% out of total M-\sys area.
Due to the absence of reported results in \aaa, we calculated above results in Table~\ref{table:comparison} given the frequency and power numbers obtained from \cite{a3:hpca20}.
The prior arts considered the scenario of enough on-chip memory with minimal consideration of the dram access cost. 
On the other hand, \sys includes all the costs from the frequent main memory access assuming the limited on-chip memory scenario by considering > 4$\times$ longer sequences (up to 4096) than prior arts.

Compared to prior work, M-\sys yields the best GOPs~/~s and GOPs~/~s~/~mm$^\mathrm{2}$ even including the main memory access cost due to it's in-memory pruning.
Compared to A$^\mathrm{3}$, M-\sys achieves 3.5$\times$ and 3.9$\times$ improvements in GOPs~/~s and GOPs~/~s~/~mm$^\mathrm{2}$ respectively.
However, it achieves 5.2$\times$ lower GOPs~/~J. This is due to two reasons: (1) the DRAM access read and write costs are not considered in the results of A$^\mathrm{3}$ and (2) the lower process technology (40 nm) in A$^\mathrm{3}$.
Taking into account the difference in the process technology node (65\,nm vs. 45\,nm), GOPs~/~J of \sys increase to 3873.5 with Dennard scaling~\cite{dennard_scaling} (1.2$\times$ lower than A$^\mathrm{3}$).
Moreover, M-\sys achieves 3.2$\times$ higher GOPs~/~s and 5.9$\times$ higher GOPs~/~s~/~mm$^\mathrm{2}$ than LeOPArd. Although the DRAM access costs are not incorporated in LeOPArd, M-\sys still delivers 1.7$\times$ higher GOPs~/~J.
Finally, M-\sys achieves 5.0$\times$, 2.4$\times$, and 4.1$\times$ enhancements in GOPs~/~s, GOPs~/~J, and GOPs~/~s~/~mm$^\mathrm{2}$, respectively, as compared to SpAtten. 
The benefits that are gained from GOPs~/~s and GOPs~/~s~/~mm$^\mathrm{2}$ are based on the  early stage in-memory pruning by leveraging the spatial locality.

Table~\ref{table:comparison} also compares GOPs~/~s~/~J~/~mm$^\mathrm{2}$ between M-\sys and prior work~\cite{a3:hpca20,spatten:hpca21,leopard:isca:2022}.
M-\sys yields 1.9$\times$ and 3.9$\times$ higher GOPs~/~s~/~J~/~mm$^\mathrm{2}$ compared to SpAtten and LeOPArd, respectively.
However, M-\sys delivers 4.8$\times$ lower GOPs~/~s~/~J~/~mm$^\mathrm{2}$ compared to \aaa, mainly due to considering DRAM access read and write costs and lower process technology node.
With Dennard scaling~\cite{dennard_scaling}, GOPs~/~s~/~J~/~mm$^\mathrm{2}$ of \sys increase to 8648.5 (3.8$\times$ better than A$^\mathrm{3}$).

\niparagraph{\rev{End-to-End comparison including fully-connected networks (FFNs).}}
\label{para:ffn}
Although SPRINT focuses on accelerating self-attention layers, the proposed accelerator can be repurposed to perform the FFN by exploiting \code{QK/V-PU} as two 8-bit input 64-tap dot-product engines. 
The $\mathcal{K}$/$\mathcal{V}$ buffers store 16KB weights of the FFN to provide 128 8-b weights per cycle by reusing the weights over many inputs. 
The M-\sys achieves speed and energy benefits for the end-to-end execution even in such small benchmarks (\bench{BERT-B}: 2.2$\times$ / 1.8$\times$, \bench{BERT-L}: 2.4$\times$ / 2.0$\times$ for energy saving / speedup) by avoiding futile computations for the padded region (see Section~\ref{sec:padding}), effectively reducing the iterations in FFN computations.
\bench{ViT-B} achieves only marginal benefit (1.1$\times$ / 1.0$\times$) due to the lack of padded area. 
M-\sys achieves greater benefit for larger benchmarks, e.g. 7.7$\times$ /  4.7$\times$ for \bench{Synth2}.

\begin{figure}[t]
\centering
\includegraphics[width=0.99\columnwidth]{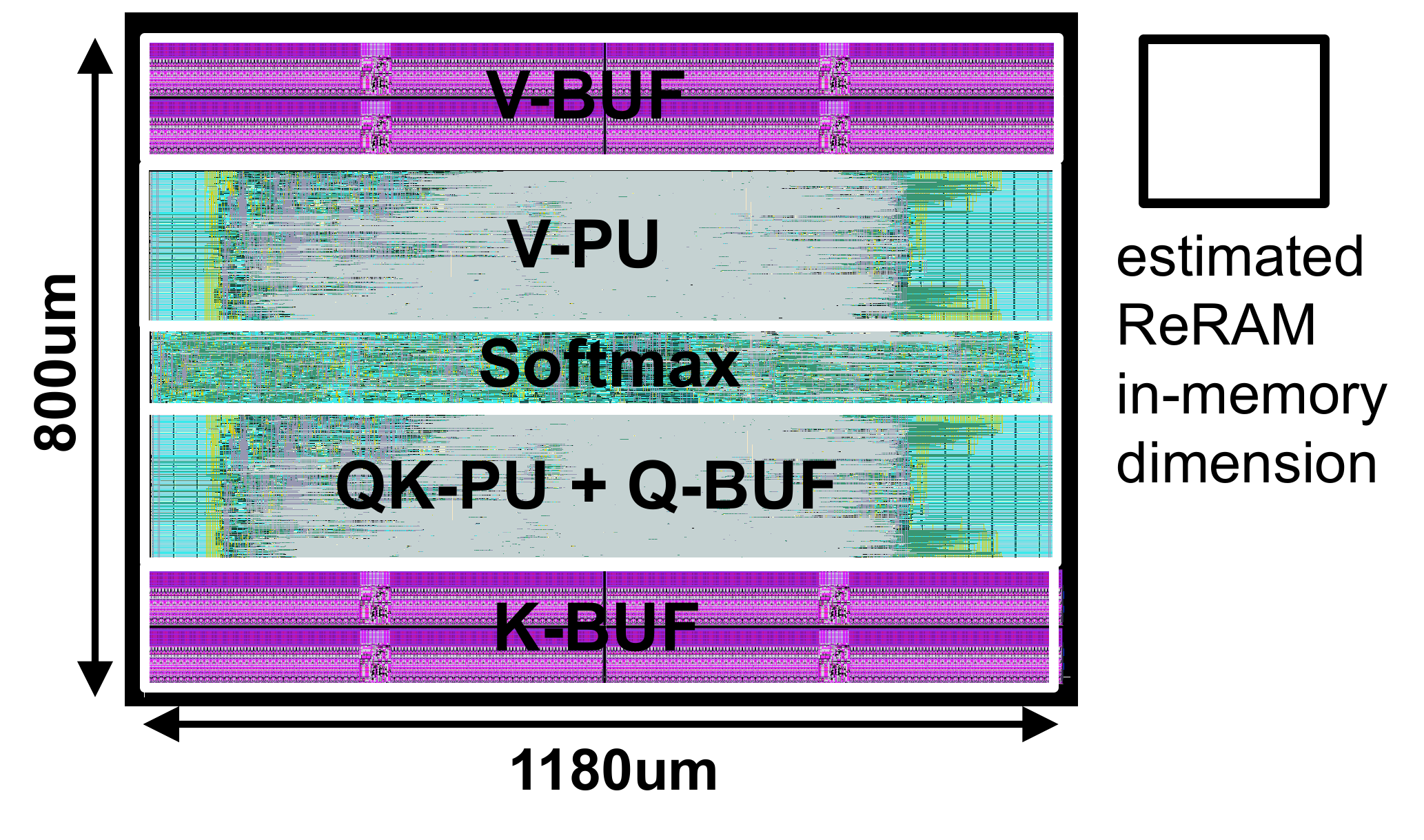}
\caption{S-\sys on-chip accelerator layout with estimated ReRAM in-memory area overhead~\cite{trans:isscc:2020}.}
\label{fig:layout} 
\end{figure}

\niparagraph{\sys on-chip accelerator and ReRAM in-memory Area.}
Figure~\ref{fig:layout} shows the  S-\sys layout in a 65~nm process which occupies $1.18\times0.8~ \mathrm{mm^2}$ including \rev{16KB on-chip SRAM}.
The layout estimation of ReRAM in-memory\rev{~\cite{trans:isscc:2020}}, including 64$\times$128 transposable array and other peripheral circuitry, is also shown in Figure~\ref{fig:layout}. 
Due to the inherent high-density of ReRAM, the area overhead takes only around \ReRAMAreaOverhead~in S-\sys.
\section{Related Work}
\label{sec:related}
Contrary to the broad spectrum of in-memory computing~\cite{execube:ipp:1994,miss:isca:1996,iram:micro:1997,flexram:iccd:1999,smartmem:isca:2000,fafnir:hpca:2021,gradpim:hpca:2021}, \sys principally positions itself as a joint in-memory analog pruning and on-chip digital recomputation system for attention-based models.
This synergistic method yields substantial gains and curtails the costly on-chip memory requirement.
These gains are maintained while preserving the baseline model accuracy.
We review the relevant literature here.

\niparagraph{In-memory computing and specialized memory controller design.}
We can broadly categorize in-memory computing into (1) 3D stacking~\cite{pimtrain:micro:2018,scalable:isca:2015,pimins:isca:2015,ndc:ispass:2014,toppim:hpdc:2014,graphp:hpca:2018,graphq:micro:2019}, (2) exploiting the inherent massive parallelism inside memory~\cite{redram:iccad:2019,computedram:micro:2019,bufcomp:date:2016,bufcomp:date:2016,rowclone:micro:2013,ambit:micro:2017,gathscatt:micro:2015,axram:pact:2018,gradpim:hpca:2021,lergan:micro:2018,fafnir:hpca:2021}, and (3) emerging memory technologies and DRAM modifications~\cite{dracc:dac:2018,xnorpop:islped:2017,drisa:micro:2017,mcdram:tcad:2018,cc:hpca:2017,xcel:tcs:2019,xcel:tcs:2019,patpim:icassp:2014,xnorsram:jssc:2020,dima:iccad:2018,parapim:dac:2019,prime:isca:2016,acdimm:isca:2013,floatpim:isca:2019,pinatubo:dac:2016,isaac:isca:2016,memris:tcs:2014,tensordimm:micro:2019,edgereram:arxiv:2021,trans:isscc:2020,reramsearch:ieee:2019,parallelreram:jetc:2022,revamp:date:2017}.
Similar to this prior work, \sys also exploits the internal structure of memory to enable a form of in-memory computation.
However, our work distinguishes itself by seamlessly blending lightweight in-memory analog approximate computing and on-chip precise recompute.
The first phase informs the on-chip accelerator to only fetch a few relevant key vectors from memory, reducing the hefty cost of data communication, while the second phase ensures model accuracy on par with baseline models.
\sys also intersects with~\cite{simulate:ispass:2014,pardis:ssca:2012,smartmc:hpca:1999,dream:ms:2016,perm:iiswc:2017,newton:micro:2020,hwsw:isca:2021,predator:codes:2007,mcslides:2018} as it similarly equips the memory controller with custom hardware blocks and specialized memory commands to unlock the full potential of in-memory thresholding.

\niparagraph{Machine learning acceleration.}
There is an abundance of prior work on accelerators for machine learning~\cite{leopard:isca:2022,yazdanbakhsh2021evaluation,bihiwe:pact20,bpvec:dac20,sigma:hpca20,planaria:micro20,bit-tactical:asplos19,encodeep,bitblade:dac19,laconic:isca19,fastwave:iccad19,tangram:asplos19,eyerissv2:journal19,simba:micro19,maestro:micro19,maeri:asplos18,brainwave:isac:2018,bitfusion:isca18,flexigan:fccm18,deepfense:iccad18,unpu:isscc:2018,ganax:isca:2018,loom:dac18,neuralcache:isca18,scnn:isca:2017,bit-pragmatic:micro17,pipelayer:hpca:2017,tetris:asplos:2017,tpu:isca:2017,eie:isca:2016,stripes:micro:2016,minerva:isca:2016,neurocube:isca:2016,isaac:isca:2016,prime:isca:2016,cambricon-x:micro:2016,cnvlutin:isca:2016,cambricon:isca:2016,eyeriss:isca:2016,dnnweaver:micro:2016,yazdanbakhsh2015neural,anpu,dadiannao:micro:2014,elsa:isca21,spatten:hpca21,sanger:micro21,edgebert:micro21,softermax:dac21,a3:hpca20,optimus:mlsys20,gobo:micro20}. 
\sys explores a different design point by seamlessly combining in-memory thresholding and on-chip recomputation to reduce the costly data communication overhead.
In addition, there is a line of work on software-only techniques that statically induce sparsity in self-attention ~\cite{rounting-transformer,longformer:2020,ye2021tr,reformer:2020,blockwise,bp-transformer, child2019generating,michel2019sixteen,wang2019structured,wen2017learning}. 
On the other hand, recent work~\cite{explicit,adaptively_sparse,fine-tune-bert} unlocks dynamic sparsity in self-attention models, yet provoking the entire computation of $\mathcal{Q}\times\mathcal{K}^T$.
%
%
Finally, a class of hardware-software methods targets early compute termination~\cite{zap:eccv20,snapea:isca:2018,compend:ics18,predictivenet:iscas17}.
While this work is not closely related, our contribution of disseminating the computations between in-memory and on-chip can be employed to perform in-memory identification of early compute opportunities.
\section{Conclusion}
\label{sec:conclusion}
Self-attention mechanisms have become integral to transformer models in multiple applications, ranging from natural language processing to computer vision. 
Despite their benefits, attention mechanisms require extravagant compute and storage space resources, quadratically proportional to input sequence length.
Recent work has presented the benefits of runtime pruning in self-attention mechanisms, albeit overlooked quadratic complexity and on-chip memory capacity requirements.
\sys harnesses the inherent parallelism of ReRAM crossbar arrays to compute the attention scores in a low-precision format.
The resulting attention scores cross a lightweight analog thresholding circuitry, which dynamically prunes the inconsequential scores.
Hence, \sys fetches only a small subset of relevant data to on-chip memory.
To mitigate the negative repercussion of approximate ReRAM computations on model accuracy, \sys recomputes the sparse attention scores for the few fetched data in digital. 
Furthermore, the paper identified and exploited a dynamic spatial locality between the adjacent attention operations even after runtime pruning.
This spatial locality further reduces the redundant data fetches and scales down the on-chip memory demand.
The combined in-memory pruning and on-chip recomputation of the relevant attention scores reduce the quadratic complexity of self-attention mechanism into a merely linear one.
The proposed shift in compute and space complexity yields significant performance gains as well as enables the acceleration of futuristic models with significantly larger input sequence length. 
\section*{Acknowledgment}
We would like to extend our gratitude towards Hadi Esmaeilzadeh, Soroush Ghodrati, Stella Aslibekyan, Suvinay Subramanian, James Laudon, and extended Google Research, Brain Team for their invaluable feedback and comments.
\section*{}
\bibliographystyle{IEEEtranS}
{\footnotesize
\bibliography{refs}}
\end{document}